\documentclass[letterpaper]{article} 
\usepackage{aaai2026}  
\usepackage{times}  
\usepackage{helvet}  
\usepackage{courier}  
\usepackage[hyphens]{url}  
\usepackage{graphicx} 
\usepackage{natbib}  
\usepackage{caption} 
\usepackage{algorithm}
\usepackage{algorithmic}

\usepackage{newfloat}
\usepackage{listings}
\DeclareCaptionStyle{ruled}{labelfont=normalfont,labelsep=colon,strut=off} 
\floatstyle{ruled}
\newfloat{listing}{tb}{lst}{}
\floatname{listing}{Listing}
\usepackage{amsmath, amssymb}

\usepackage{booktabs} 
\usepackage{multirow} 
\usepackage{tabularx}
\usepackage{arydshln} 

\usepackage{tcolorbox}
\tcbuselibrary{skins, breakable}
\usepackage{xcolor}
\usepackage{lipsum}
\usepackage{etoolbox} 

\title{Textual Self-Attention Network: Test-Time Preference Optimization Through Textual Gradient-Based Attention}

\author{
   Shibing Mo\textsuperscript{\rm 1}\textsuperscript{\rm 2}\equalcontrib, 
   Haoyang Ruan\textsuperscript{\rm 1}\equalcontrib,
   Kai Wu,\textsuperscript{\rm 1}\thanks{Corresponding author.}
   Jing Liu\textsuperscript{\rm 1}\textsuperscript{\rm 2}
}
\affiliations{
    \textsuperscript{\rm 1}School of Artifcial Intelligence, Xidian University\\
    \textsuperscript{\rm 2}Guangzhou Institute of Technology, Xidian University\\

    kwu@xidian.edu.cn
}

\usepackage{bibentry}

\begin{document}

\maketitle

\begin{abstract}
Large Language Models (LLMs) have demonstrated remarkable generalization capabilities, but aligning their outputs with human preferences typically requires expensive supervised fine-tuning. Recent test-time methods leverage textual feedback to overcome this, but they often critique and revise a single candidate response, lacking a principled mechanism to systematically analyze, weigh, and synthesize the strengths of multiple promising candidates. Such a mechanism is crucial because different responses may excel in distinct aspects (e.g., clarity, factual accuracy, or tone), and combining their best elements may produce a far superior outcome. This paper proposes the Textual Self-Attention Network (TSAN), a new paradigm for test-time preference optimization that requires no parameter updates. TSAN emulates self-attention entirely in natural language to overcome this gap: it analyzes multiple candidates by formatting them into textual keys and values, weighs their relevance using an LLM-based attention module, and synthesizes their strengths into a new, preference-aligned response under the guidance of the learned textual attention. This entire process operates in a textual gradient space, enabling iterative and interpretable optimization. Empirical evaluations demonstrate that with just three test-time iterations on a base SFT model, TSAN outperforms supervised models like Llama-3.1-70B-Instruct and surpasses the current state-of-the-art test-time alignment method by effectively leveraging multiple candidate solutions.

\end{abstract}

\begin{links}
    \link{Code}{https://github.com/Explorermomo/TSAN-main}
    \link{Extended version}{http://arxiv.org/abs/2511.06682}
\end{links}


\section{Introduction}

Aligning LLMs with human values and preferences has become a cornerstone of modern AI research, ensuring these powerful systems are helpful, harmless, and honest \cite{1}. The dominant paradigms for achieving this alignment, such as reinforcement learning from human feedback (RLHF) \cite{2} and direct preference optimization (DPO) \cite{3}, are performed during the training phase \cite{4}. These methods embed preferences directly into the model's parameters, producing statically-aligned AI models. While effective, this approach is computationally intensive and lacks the flexibility to adapt to new or evolving preferences without expensive retraining cycles \cite{5}.

\begin{figure}[t]
\centering
\includegraphics[width=1.0\columnwidth]{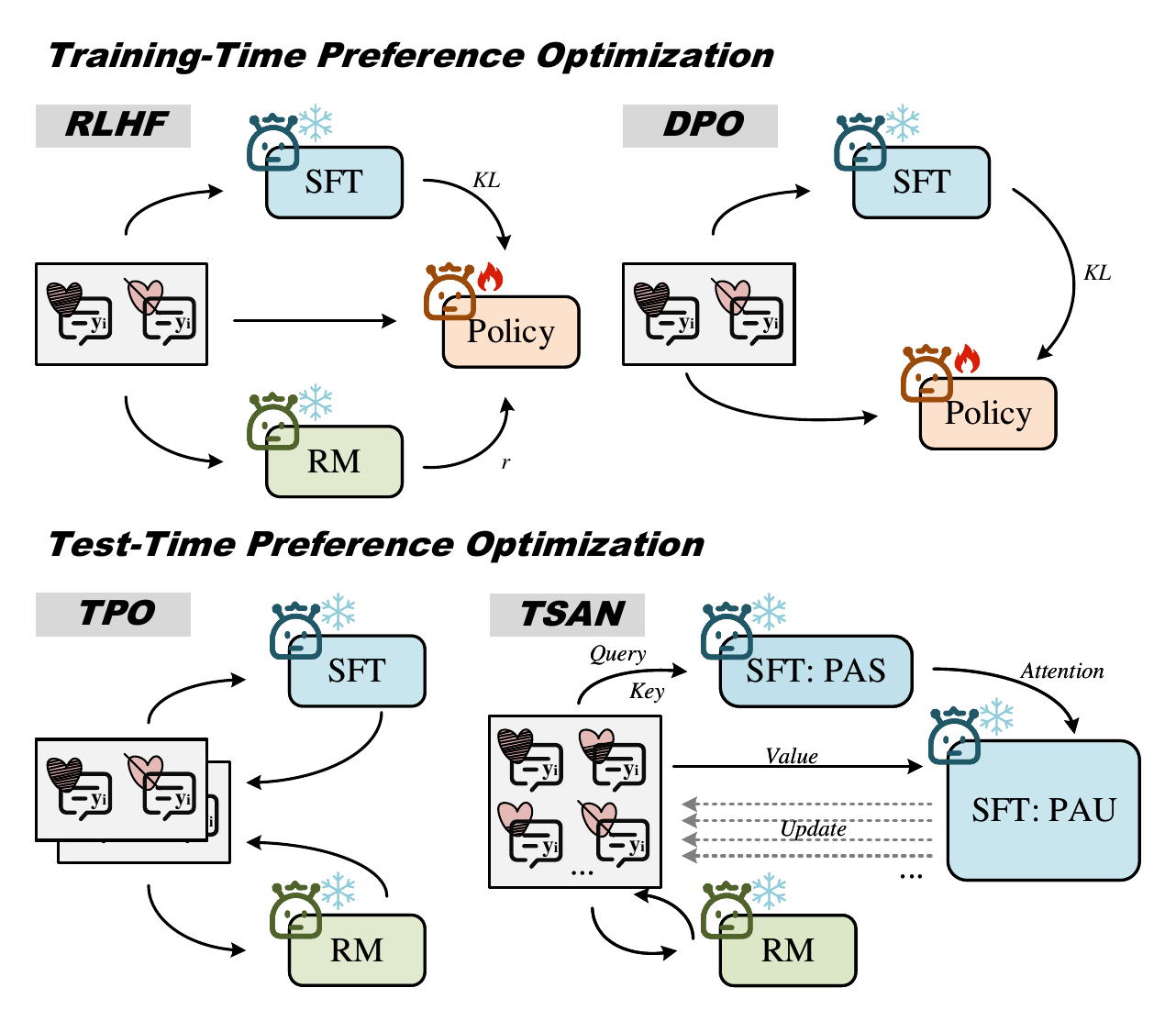} 
\caption{Comparison of Textual Self-Attention Network with existing preference optimization methods (e.g., RLFH, DPO, and TPO), where PAS and PAU represent textual attention scores and textual aggregation updates, respectively. }
\label{fig1}
\end{figure}

This inflexibility has given rise to an emerging area of test-time self-improvement \cite{6}. These methods seek to align LLM outputs on-the-fly, trading additional test-time computation for higher-quality, better-aligned responses during the generation process \cite{7}. However, many existing test-time preference optimization techniques face significant limitations. Rudimentary methods like Best-of-N sampling \cite{8} rely on a simple scalar reward signal to select the best response from a pool of candidates. This scalar reward is a notorious information bottleneck and is highly susceptible to reward hacking, where the model learns to exploit proxy signals (e.g., generating verbose or formulaic answers) to achieve a high score without genuinely improving quality \cite{9}.



More sophisticated approaches have begun to leverage richer, textual feedback to overcome the limitations of scalar rewards. Frameworks like test-time preference optimization  (TPO) \cite{10} and critique $\&$ revise \cite{11} generate natural language critiques and use them as textual gradients to iteratively improve a single response. As shown in figure \ref{fig1}, although this is an important step forward, a key limitation remains: these methods still operate in a linear, unstructured fashion. They typically critique one candidate response and then attempt to revise it, lacking a principled mechanism to systematically analyze, weigh, and synthesize the strengths of multiple promising candidate outputs simultaneously. This single-path revision process misses the opportunity to combine the best aspects of several strong alternatives, fundamentally limiting its optimization potential.

To address this gap, we introduce the TSAN, a novel test-time preference optimization method that operationalizes the principles of a self-attention mechanism within the textual domain. Rather than merely critiquing and revising a single output, TSAN formalizes a structured process for aggregating information from a set of high-quality candidate responses. It treats the user's prompt as a query (Q) and a curated set of strong initial answers as textual keys (K) and values (V). The framework then uses an LLM to generate textual attention scores — a natural language analysis of each key's relevance and merit with respect to the query. Guided by this rich, contextual attention signal, TSAN synthesizes a new, superior response by performing an aggregation update that integrates the most salient features from the Values. This entire process is enclosed in an iterative optimization loop driven by textual gradient descent, allowing the model to progressively refine its output for better alignment with user preferences. In summary, TSAN distinguishes itself from previous works through some key features:
\begin{itemize}
    \item We introduce a new paradigm for test-time alignment focused on the principled, compositional synthesis of an optimal response from multiple candidates. This moves beyond the prevailing approach of selecting or revising a single response, establishing a more structured method to systematically analyze and integrate the strengths of diverse solutions.
    \item To realize this paradigm, we propose the TSAN, an innovative framework that operationalizes a query-key-value self-attention mechanism entirely within the textual domain. By generating natural language attention scores to weigh the merits of each candidate, TSAN effectively processes multiple inputs to synthesize a superior, preference-aligned output without any parameter updates.
    \item Extensive experiments have demonstrated that by structurally designing the refinement process through a textual attention mechanism, TSAN can provide a more robust and effective test-time alignment method.
\end{itemize}

\section{Related Work}
This work is situated at the intersection of LLM preference optimization, test-time alignment, and the emerging field of textual feedback mechanisms.

\subsection{Preference Optimization Paradigms}
Research in LLM alignment is broadly divided into two paradigms. Training-time alignment methods instill preferences by modifying model weights. This began with the complex, multi-stage pipeline of RLHF, which involves training a separate reward model and using reinforcement learning (like PPO \cite{12}) to optimize the policy. Due to its instability and high computational cost, the field has largely shifted towards simpler and more stable methods like DPO \cite{3}. DPO and its successors, such as RLOO \cite{14}, CPO \cite{15}, and ORPO \cite{16}, reframe the preference learning problem as a classification task, thereby eliminating the need for an explicit reward model and a complex reinforcement learning loop. However, all these methods produce a static model that cannot be adjusted post-training.

In contrast, test-time alignment aims to provide this missing flexibility by intervening during the generation process without updating model weights. These techniques range from decoding strategies and sampling methods to more complex iterative refinement loops. As shown in figure \ref{fig1}, TSAN falls squarely into this paradigm, offering a novel mechanism for on-the-fly alignment.

\subsection{Test-Time Self-Improvement}

The field of test-time self-improvement has explored several strategies for enhancing model outputs at test time.

\subsubsection{Sampling and Search.} The most straightforward approach is Best-of-N sampling \cite{8}, where N candidate responses are generated, and the one with the highest score from a reward model is selected. More advanced techniques, such as Tree-of-Thoughts \cite{17} and TreeBoN \cite{18}, employ tree-based search to explore the solution space more effectively. However, these methods typically rely on scalar rewards and are thus vulnerable to reward hacking.

\subsubsection{Iterative Refinement.} For more nuanced correction, iterative methods have been proposed. The critique $\&$ revise framework involves generating an initial response, critiquing it, and then generating a revision based on the critique \cite{10,11}. TPO formalizes this concept by treating the critique as the textual gradient \cite{19} that guides optimization. As shown in figure \ref{fig1}, TSAN builds upon this iterative foundation but introduces a key distinction: instead of a linear loop of critiquing and revising a single candidate, TSAN implements a structured, attention-based mechanism to synthesize an improved response from a whole set of strong candidates. More related work can be found in \textbf{Appendix A}.

\section{Preliminary}

The goal of preference optimization in LLMs is to align the output of a policy $\pi_{\theta}$ with human preferences. This objective is typically formalized as maximizing a scoring function $s$ over a preference dataset $D$,  which consists of a prompt $x$, a chosen response $y_w$, and a rejected response $y_l$:

\begin{equation}
    \max_{\pi} E_{(x,y_w,y_l) \sim \mathcal{D}} \big[ s(\pi, x, y_w, y_l) \big]    
\end{equation}

\subsection{Training-Time Preference Optimization}
Mainstream alignment methods achieve this during the training phase by updating the model parameters $\theta$. RLHF is a seminal work in this area. It first trains a reward model (RM), $r_\phi$, to fit human preferences. Then, it uses reinforcement learning to maximize this reward for the policy model, while a KL-divergence term penalizes its deviation from a reference model $\pi_{\text{ref}}$:
\begin{equation}
\max_{\theta} \mathbb{E}_{x \sim \mathcal{D}, y \sim \pi_\theta(y|x)} \big[ r_\phi(x, y) - \beta \cdot KL\big(\pi_\theta(y|x) \,\big|\big|\, \pi_{\text{ref}}(y|x)\big) \big]
\end{equation}

\subsection{Test-Time Preference Optimization}

In contrast to methods that update model weights, test-time optimization intervenes on the model's output during the inference phase. An emerging direction in this area is the use of textual gradients. This paradigm does not compute numerical gradients $\nabla_\theta L(\theta)$. Instead, it leverages the LLMs' own capabilities to iteratively revise its output text $y$ guided by natural language critiques. This process can be abstracted as:
\begin{equation}
y_{i+1} \leftarrow \text{Revise}(y_i, \nabla_{\text{text}})~, \quad \text{where}~ \nabla_{\text{text}} = \text{Critique}(y_i, L_{\text{text}})
\end{equation}
The TSAN we propose provides a novel and structured framework for applying textual gradients under this test-time optimization paradigm.

\section{Method}

To address the insufficiency of existing test-time preference optimization methods in integrating information from multiple candidate answers, we propose the TSAN. TSAN is to elevate the self-attention mechanism from the numerical level to the textual level at test time. Through a structured, LLM-driven process, it enables the systematic analysis, weighting, and synthesis of multiple high-quality candidate answers, thereby generating a more optimal response.

The overall framework of TSAN comprises three core phases: (1) Candidate generation and textual QKV construction; (2) Textual attention calculation; (3) Aggregate and update with alternating optimization. The entire process operates within an optimization loop driven by textual gradients.

\subsection{Candidate Generation and Textual QKV Construction}

Given a user input $x$, the TSAN process is initiated.

\begin{enumerate}
    \item \textbf{Candidate Generation}: TSAE first utilize a policy model $\pi_\theta$ to generate a set of $N$ diverse candidate responses $\{y_1, y_2, \ldots, y_N\}$ for the input $x$. This step initially explores the possible solution space.
    \item \textbf{Reward Model Scoring and Selection}: Next, using an independent RM to score these $N$ candidates, obtaining their respective preference scores. Then cache each candidate response and its score, and sort them in descending order based on their scores:
    \begin{equation}
    Y = \{(y_i, R(y_i))\}_{i=1}^N
    \end{equation}
    \item \textbf{Textual QKV Construction}: Inspired by the self-attention mechanism, we reframe the optimization problem as an interaction of query, key, and value in the textual domain.
    \begin{enumerate}
        \item \textbf{Textual Query (Q)}: The user's original input $x$ is directly used as the textual query, $Q_{\text{text}}$.
        \item \textbf{Textual Key (K) and Value (V)}: Selecting the top-$k$ candidate responses from the cache and concatenate them into a single, numbered string. This string serves simultaneously as the textual key, $K_{\text{test}}$, and the textual value, $V_{\text{test}}$. In TSAN, $K_{\text{test}}$ is used to evaluate the relevance of each candidate answer to the query, while $V_{\text{test}}$ acts as the carrier for the content of these answers, which will ultimately be synthesized. Formally:
        \begin{equation}
        Q_{\text{text}} = x
        \end{equation}
        \begin{equation}
        K_{\text{test}} = V_{\text{test}} = \text{Top-}k\big(\{(y_i, R(y_i))\}_{i=1}^N\big)
        \end{equation}
    \end{enumerate}
\end{enumerate}

\subsection{Textual Attention Score}

In standard self-attention mechanisms, attention scores are numerical matrices calculated by computing the dot product of query vectors and key vectors. However, in TSAN, we use specially configured LLMs to generate a natural language format analysis report, which we refer to as \textbf{textual attention scores} ($AS_{text}$).

Specifically, we construct an attention score model, the $\text{PAS}_{\text{model}}$, which is an LLM configured with a specific system prompt $\text{P}_{\text{att}}$. This prompt instructs the model to role-play as an attention mechanism, with the following task:

\begin{quote}
"\textit{Analyze the relevance between the query $Q$ and each candidate key $K_i$. Evaluate the key characteristics of each $K_i$ regarding aspects such as performance and accuracy, and provide a step-by-step explanation. Finally, perform a qualitative summary of these characteristics to produce a coherent textual description that synthesizes all insights.}"
\end{quote}

By feeding the textual query $Q_{\text{text}}$ and the textual key $K_{\text{text}}$ as input into this model, its output is the textual attention score, $AS_{\text{text}}$. This $AS_{\text{text}}$ is a piece of text rich with associative information; it not only assesses relevance but also points out the advantages and disadvantages of each candidate answer, providing high-quality, interpretable guiding signals for the subsequent aggregation step.

\begin{equation}
AS_{\text{text}} = \text{PAS}_{\text{model}}(Q_{\text{text}}, K_{\text{text}})
\end{equation}

\subsection{Aggregate Update}

After obtaining the textual attention score, the next step is to utilize this score to guide a model in synthesizing the merits of all candidate answers (i.e., the textual value, $V_{\text{text}}$) to generate a completely new and more optimal response. We call this process \textbf{textual aggregate update}.

Similar to the previous step, we construct an \textbf{aggregate update model } $\text{PAU}_{\text{model}}$, which is configured by a system prompt $\text{P}_{\text{agg}}$. This prompt instructs the model:

\begin{quote}
"\textit{You are an aggregation and update mechanism. Your task is, based on the textual attention score ($AS_{\text{text}}$) that analyzes the relationship between Q and each $K_i$, and in conjunction with the corresponding $V_i$, to integrate the information contained within these values and synthesize a new, rich output.}"
\end{quote}

The inputs $Q_{\text{text}}$, $AS_{\text{text}}$, and $V_{\text{text}}$ are fed into this model. The model's output, a synthetically refined answer, is then cached and denoted as $y_{\text{agg}}$. Conceptually, this step is equivalent to the process in self-attention where attention weights are applied to the value vectors, which are then summed to produce an output. In TSAN, however, this is an inference process based entirely on textual understanding and generation.

\begin{equation}
y_{\text{agg}} = \text{PAU}_{\text{model}}(Q_{\text{text}}, AS_{\text{text}}, V_{\text{text}})
\end{equation}

\begin{equation}
Y = \left\{(y_j, R(y_j))\right\}_{j=1}^N \cup y_{\text{agg}}
\end{equation}

\subsection{Iterative Optimization}

During the test-time optimization process, the preceding steps are embedded into an iterative optimization loop. Driven by textual gradients, the system continuously adjusts the output to better align with the preferences of the reward model. This process is composed of the following parts:


\begin{table*}[!t]
    \centering

    \begin{tabular}{lccccccc}
        \toprule
        \multirow{2}{*}{Model} & \multicolumn{2}{c}{AlpacaEval 2} & \multirow{2}{*}{Arena-Hard 2} & 
        \multirow{2}{*}{HH-RLHF} & \multirow{2}{*}{BeaverTails} & \multirow{2}{*}{XSTest} & \multirow{2}{*}{MATH-500} \\
         & LC(\%) & WR(\%)  \\
        \midrule
        Llama-3.1-70B-DPO (IQ4) & 9.47 & 15.79 & 6.3 & -2.94 & -7.30 & 64.1 & 25.8 \\
        Llama-3.1-70B-Instruct (IQ4) & \textbf{21.73} & 18.18 & 6.8 & \textbf{-2.45} & -7.50 & 69.2 & 24.0 \\
        Llama-3.1-70B-SFT (IQ4) & 3.01 & 4.91 & 5.5 & -6.65 & -10.07 & 75.2 & 22.0 \\
        \addlinespace[0.18em]
        \cdashline{1-8}[5pt/3pt]
        \addlinespace[0.4em]
        SFT - TPO (k4-T3) & 17.95 & \underline{20.18} & 6.0 & -2.96 & -6.67 & \underline{76.6} & \textbf{32.0} \\
        SFT - TPO (k4-T5) & 18.28 & \textbf{22.15} & 6.0 & -2.96 & \underline{-6.53} & \underline{76.6} & \textbf{32.0} \\
        SFT - TSAN (k4-M4-T3) & \underline{18.57} & 17.05 & \underline{8.4} & \underline{-2.88} & -6.63 & \textbf{78.8} & \underline{28.2} \\
        SFT - TSAN (k4-M4-T5) & \underline{18.57} & 17.05 & \textbf{8.5} & \underline{-2.88} & \textbf{-6.45} & \textbf{78.8} & \underline{28.2} \\
        \bottomrule
    \end{tabular}
    \caption{Unaligned Model Performance Comparison on Multiple Benchmarks, where the IQ4 indicates INT4 quantization, k indicates the number of candidate answers sampled, M indicates the number of textual attention heads, and T indicates the number of TSAN iterations. The \textbf{bold} and  \underline{underlined} numbers indicate the best and second-best performances, respectively. }
    \label{tab : unaligned-perf}
\end{table*}

\begin{enumerate}
    \item \textbf{Textual Loss Calculation}: We use a prompt $P_l$ to express the loss function, where an $LLM_L$ is prompted to critique the previously generated answers \( \{y_{\text{agg}}^i\}_{i=1}^M \) with respect to the original user prompt \( x \). The process can be formalized as:
    \begin{equation}
    L_{\text{text}} = LLM_L(P_l, x, y_{\text{agg}})
    \end{equation}
    where the prompt $P_l$ instructs \( \text{LLM}_L \):
    \begin{quote}
        "\textit{You are supposed to evaluate the model's response. Please assess the strengths and weaknesses of the model's answer,then refine the PAS\_prompt to instruct the PAS\_Model to focus more on certain aspects and ensure better alignment of attention,and refine the PAU\_prompt to guide the PAU\_Model on how to utilize the attention analysis results from the PAS\_Model to generate a better answer from the existing answers. Step by step.}"
    \end{quote}
    
    It can be seen that it not only to evaluate \( y_{\text{agg}} \) but also to provide feedback for improving the TSAN framework itself. Therefore, the resulting textual loss, \( L_{\text{text}} \), is a structured critique containing feedback for various learnable components of the framework. We can formally represent this output as a tuple:
    \begin{equation}
    L_{\text{text}} = (\nabla_{y_\text{agg}}, \nabla_{P_\text{att}},\nabla_{P_\text{agg}})
    \end{equation}

    \item \textbf{Gradient Computation}: A prompt \( P_{\text{grad}} \) is used to transform the textual loss \( L_{\text{text}} \) described above into an update instruction, thereby forming a textual gradient:
    \begin{equation}
    \frac{\partial \mathcal{L}}{\partial y_{\text{agg}}} = LLM(P_{\text{grad}}( L_{\text{text}}))
    \end{equation}

    \item \textbf{Variable Optimization}: Finally, a prompt \( P_{\text{update}} \) is used to update the variables to generate an optimized set \( \{{y_{\text{agg}}^i}'\}_{i=1}^M \), analogous to a gradient descent update rule:
    \begin{equation}
    \{{y_{\text{agg}}^i}'\}_{i=1}^M = LLM\left(P_{\text{update}}( \frac{\partial \mathcal{L}}{\partial y_{\text{agg}}})\right)
    \end{equation}

    where the \( M \) parallel outputs can be considered as the outputs of \( M \) attention heads. These newly generated answers are re-scored by the reward model and are used to update the textual key \( K_{\text{text}} \) and textual value \( V_{\text{text}} \) for the next iteration.
    \begin{equation}
    Y = \left\{(y_j, R(y_j))\right\}_{j=1}^N \cup y_{agg} \cup \left\{{y_{\text{agg}}^i}'\right\}_{i=1}^M
    \end{equation}
\end{enumerate}

The entire process (from textual attention score calculation to the optimizer steps) is repeated until a preset maximum number of iterations, \( \text{max\_iters} \), is reached. Besides, We adopt the vanilla prompts for $P_{grad}$ and $P_{update}$ from TextGrad \cite{19}. The complete steps are detailed in \textbf{Algorithm \ref{alg:tsan}}. 

\begin{algorithm}[t!]
    \caption{Textual Self-Attention Network (TSAN)}
    \label{alg:tsan}
    \begin{algorithmic}[1]
        \REQUIRE Query $x$, policy model $\pi_{\theta}$, reward model RM,  max iterations $T$
        \ENSURE The final optimized answer $y_{final}$
        \STATE // Initialization
        \STATE $Y_{cand} \leftarrow$ Generate $N$ candidate responses $\{y_1, \dots, y_N\}$ in parallel using $\pi_{\theta}$.
        \STATE $K_{text}^{(0)}, V_{text}^{(0)} \leftarrow$ Evaluate $Y_{cand}$ using RM, select the Top-k, and concatenate into a string.
        \STATE $Q_{text}  \leftarrow x$
        \FOR{$i = 0$ to $T-1$}
            \STATE // Textual Attention Calculation \\
                 $AS_{text}^{(i)} \leftarrow \text{PAS\_model}(Q_{text}, K_{text}^{(i)})$
            
            \STATE // Aggregate and Update \\
             $y_{agg} \leftarrow \text{PAU\_model}(Q_{text}, AS_{text}^{(i)}, V_{text}^{(i)})$

            \STATE // Textual Gradient Optimization \\
             $\nabla_{y_{agg}}^{(i)} \leftarrow L_{text}(y_{agg})$ // Generate critique
            \STATE $\{{y_{\text{agg}}^i}'\}_{i=1}^M \leftarrow \text{Optimizer.step}(\nabla_{y_{agg}}^{(i)})$ // Generate new candidate set based on the critique
            \STATE // Update candidates \\
             $Y_{cand} \leftarrow  \{{y_{\text{agg}}^i}'\}_{i=1}^M \cup y_{agg} \cup Y_{cand}$
            \STATE // Update K and V for the next iteration \\
             $K_{text}^{(i+1)}, V_{text}^{(i+1)} \leftarrow$ Evaluate $Y_{cand}$ using RM and select the Top-k.
        \ENDFOR
        \STATE $y_{final} \leftarrow$ Select the highest-scoring response from all candidates generated across all iterations.
        \STATE \textbf{Return} $y_{final}$
    \end{algorithmic}
\end{algorithm}

\section{Experiments}
\subsection{Experimental Setup}

\subsubsection{Models.}

This research contrasts two policy model categories: unaligned and aligned, with the core distinction lying in whether preference optimization (e.g., RLHF or DPO) is applied. Specifically, for the unaligned model, we adopt Llama-3.1-Tulu-3-70B-SFT (IQ4 XS) \cite{e1}, obtained by supervised fine-tuning (SFT) on the foundation model Llama-3.1-70B \cite{e2}. For aligned models, we employ two publicly available models: Llama-3.1-70B-Instruct (IQ4 XS) \cite{e2} and the smaller-scale Mistral-Small-Instruct-2409 \cite{e3}. For RM, we use FsfairX-LLaMA3-RM-v0.1 \cite{e4} for all policy models. Additionally, we follow the settings of TPO \cite{10}, training an on-policy aligned model (termed Llama-3.1-70B-DPO) based on Llama-3.1Tulu-3-70B-SFT using UltraFeedback \cite{e5}. More optimize Details can be found in \textbf{Appendix B1.}

\begin{table*}[!t]
    \centering

    \begin{tabular}{lccccccc}
        \toprule
        \multirow{2}{*}{Model} & \multicolumn{2}{c}{AlpacaEval 2} & \multirow{2}{*}{Arena-Hard 2} & 
        \multirow{2}{*}{HH-RLHF} & \multirow{2}{*}{BeaverTails} & \multirow{2}{*}{XSTest} & \multirow{2}{*}{MATH-500} \\
         & LC(\%) & WR(\%)  \\
        \midrule
        Llama-3.1-70B-Instruct (IQ4) & 21.73 & 18.18 & 6.8 & -2.45 & -7.50 & 69.2 & 24.0 \\
        \addlinespace[0.18em]
        \cdashline{1-8}[5pt/3pt]
        \addlinespace[0.4em]
        + TSAN (k4-M4-T3) & \textbf{26.51} & \textbf{23.19} & 8.1 & -1.73 & -6.63 & \textbf{81.7} & \textbf{38.0} \\
        + TSAN (k4-M4-T5) & \textbf{26.51} & \textbf{23.19} & \textbf{10.4} & \textbf{-1.52} & \textbf{-6.44} & \textbf{81.7} & \textbf{38.0} \\
        \midrule
        Mistral-Small-Instruct-2409 & 15.76 & 16.10 & 9.1 & -3.11 & -6.95 & 72.0 & 27.6 \\
        \addlinespace[0.18em]
        \cdashline{1-8}[5pt/3pt]
        \addlinespace[0.4em]
        + TSAN (k4-M4-T3) & \textbf{22.57} & 24.10 & \textbf{10.3} & \textbf{-2.46} & -6.25 & \textbf{74.4} & \textbf{32.2} \\
        + TSAN (k4-M4-T5) & \textbf{22.57} & \textbf{25.38} & \textbf{10.3} & \textbf{-2.46} & \textbf{-6.13} & \textbf{74.4} & \textbf{32.2} \\
        \bottomrule
    \end{tabular}
    \caption{Aligned Model Performance Comparison on Multiple Benchmarks, where the IQ4 indicates INT4 quantization, k indicates the number of candidate answers sampled, M indicates the number of textual attention heads, T indicates the number of TSAN iterations and \textit{+ TSAN} indicates the use of TSAN based on the Llama-3.1-70B-Instruct (IQ4) model or the Mistral-Small-Instruct-2409 model. The \textbf{bold} indicate the best performances. }
    \label{tab : aligned-perf}
\end{table*}

\begin{table*}[!t]
    \centering

    \begin{tabular}{lccccccc}
        \toprule
        \multirow{2}{*}{Model} & \multicolumn{2}{c}{AlpacaEval 2} & \multirow{2}{*}{Arena-Hard 2} & 
        \multirow{2}{*}{HH-RLHF} & \multirow{2}{*}{BeaverTails} & \multirow{2}{*}{XSTest} & \multirow{2}{*}{MATH-500} \\
         & LC(\%) & WR(\%)  \\
        \midrule
        Qwen-3-Plus & \textbf{65.08} & 68.38 & 47.3 & \textbf{2.47} & -4.26 & 75.1 & 90.0 \\
        \addlinespace[0.18em]
        \cdashline{1-8}[5pt/3pt]
        \addlinespace[0.4em]
        + TSAN (k4-M4-T3) & 57.56 & 69.90 & \textbf{72.1} & 1.96 & -3.16 & \textbf{81.9} & 90.6 \\
        + TSAN (k4-M4-T5) & 57.56 & \textbf{71.69} & \textbf{72.1} & 1.96 & \textbf{-3.05} & \textbf{81.9} & \textbf{90.8} \\
        \bottomrule
    \end{tabular}
    \caption{TSAN Performance (Model Scores) on Qwen-3-Plus, k indicates the number of candidate answers sampled, M indicates the number of textual attention heads, T indicates the number of TSAN iterations and \textit{+ TSAN} indicates the use of TSAN based on the Qwen-3-Plus API. The \textbf{bold} indicate the best performances. }
    \label{tab : aligned-perf-Qwen}
\end{table*}


\begin{table*}[!t]
    \centering
    \begin{tabular}{lccccccc}
        \toprule
        \multirow{2}{*}{Model} & \multicolumn{2}{c}{AlpacaEval 2} & \multirow{2}{*}{Arena-Hard 2} & 
        \multirow{2}{*}{HH-RLHF} & \multirow{2}{*}{BeaverTails} & \multirow{2}{*}{XSTest} & \multirow{2}{*}{MATH-500} \\
         & LC(\%) & WR(\%)  \\
        \midrule
        gpt-oss 20B & 61.60 & 70.27 & 72.5 & 1.63 & -7.45 & 84.5 & 96.0 \\
        gpt-oss 20B + TPO (T5) & 63.90 & 72.70 & 75.2 & 1.44 & -6.20 & 86.0 & 95.0 \\
        gpt-oss 20B + TSAN (T5) & \textbf{67.90} & \textbf{75.90} & \textbf{82.8} & \textbf{1.68} & \textbf{-5.98} & \textbf{86.3} & \textbf{97.6} \\
        \addlinespace[0.18em]
        \cdashline{1-8}[5pt/3pt]
        \addlinespace[0.4em]
        gpt-oss 120B  & 67.70 & 76.90 & 84.4 & 2.52 & -6.54 & 85.0 & \textbf{96.4} \\
        gpt-oss 120B + TSAN (T5) & \textbf{72.60} & \textbf{78.90} & \textbf{87.4} & \textbf{2.57} & \textbf{-5.57} & \textbf{88.9} & \textbf{96.4} \\
        \bottomrule
    \end{tabular}
    \caption{TSAN Performance (Model Scores) on gpt-oss model. Configurations: k indicates the number of candidate answers sampled (k=4 for all experiments), M indicates the number of textual attention heads (M=4 for all experiments), T indicates the number of iterations (noted in the table).}
    \label{tab : gpt-oss}
\end{table*}

\subsubsection{Evaluation Benchmarks. }

We evaluate our model using a comprehensive set of benchmarks covering multiple facets, including instruction following (AlpacaEval 2 \cite{e7} and Arena-Hard 2 \cite{e8}), general preference alignment (HH-RLHF \cite{e9}), safety (BeaverTails \cite{1} and XSTest \cite{e11}), and mathematical capabilities (MATH-500 \cite{8}). More details about the dataset can be found in \textbf{Appendix B2}.

\subsection{Experimental Results}


We evaluate the effectiveness of our proposed TSAN by applying it to both unaligned and aligned base models. The results, detailed in table \ref{tab : unaligned-perf} and table \ref{tab : aligned-perf}, demonstrate that TSAN serves as a powerful and versatile test-time optimization framework, significantly enhancing model performance across a wide array of benchmarks.

\subsection{Benchmark Performance}

\subsubsection{Unaligned Models.}
As presented in table \ref{tab : unaligned-perf}, applying TSAN to the base supervised fine-tuned model (Llama-3.1-70B-SFT) yields substantial improvements, elevating its capabilities to be highly competitive with training-time alignment methods like DPO and TPO.

Most notably, on instruction-following benchmarks, TSAN brings dramatic gains. On AlpacaEval 2, it boosts the length-controlled win rate (LC) from a mere 3.01\% to 18.57\% and the raw WR from 4.91\% to 17.05\%. This demonstrates a vastly improved ability to adhere to user instructions. Similarly, on the more challenging Arena-Hard 2 benchmark, TSAN improves the score from 5.5\% to 8.5\%, surpassing the performance of the SFT-TPO model (6.0\%).

In terms of preference alignment and safety, TSAN consistently steers the SFT model towards better-aligned and safer outputs. The average reward on HH-RLHF improves from -6.65 to -2.88, and on BeaverTails from -10.07 to -6.45. This indicates that TSAN effectively guides the model to generate responses that are better aligned with the reward model's learned human preferences. On XSTest, TSAN increases the compliance accuracy from 75.2\% to 78.8\%. Furthermore, TSAN enhances the model's reasoning capabilities, as shown by the improvement on MATH-500 from 22.0\% to 28.2\%.

When compared with SFT-TPO, a strong training-time baseline, our test-time SFT-TSAN method proves to be highly competitive. While TPO achieves a higher raw win rate on AlpacaEval 2 and a stronger score on MATH-500, TSAN outperforms TPO on Arena-Hard 2, XSTest, and achieves slightly better reward scores on HH-RLHF and BeaverTails. This highlights that TSAN can achieve a comparable level of alignment to training-time methods without requiring any gradient-based updates or extensive training data. And RM scores result can be found in \textbf{figure A1} in \textbf{Appendix C}.

\subsubsection{Aligned Models.}
As shown in table \ref{tab : aligned-perf}, TSAN serves as a potent test-time enhancement for already-aligned models of varying scales. On the large-scale Llama-3.1-70B-Instruct, it delivers substantial gains across all benchmarks, notably improving the win rate on AlpacaEval 2 (WR from 18.18\% to 23.19\%) and Arena-Hard 2 (from 6.8\% to 10.4\%). The most striking improvements are in reasoning and safety, where TSAN boosts accuracy on MATH-500 from 24.0\% to a remarkable 38.0\% and on XSTest from 69.2\% to 81.7\%.

To demonstrate the generality of our method, we also applied TSAN to the smaller-scale Mistral-Small-Instruct-2409 model. A consistent pattern of significant improvement emerges. TSAN boosts Mistral-Small's performance on AlpacaEval 2 (WR from 16.10\% to 25.38\%) and Arena-Hard 2 (from 9.1 to 10.3), while also enhancing its MATH score to 32.2\%. This demonstrates that TSAN is a model-agnostic enhancement technique, capable of delivering substantial performance lifts across different model architectures and scales.

Furthermore, increasing iterations from T3 to T5 consistently yields better results for both models, especially on Arena-Hard 2 and in reward scores on HH-RLHF, confirming the value of continued optimization. In summary, TSAN proves to be a versatile post-hoc enhancement method, unlocking significant performance gains on SOTA aligned models without any additional training. And RM scores result can be found in \textbf{Appendix C}. 



\subsubsection{TSAN Performance on Qwen-3-Plus.}

Experimental results in table \ref{tab : aligned-perf-Qwen} show that a key advantage of TSAN is its exceptional portability. Because the entire optimization process operates in the textual domain without requiring access to model weights or gradients, it can be applied to powerful, closed-source models as a "plug-and-play" enhancement through simple API calls.

To validate this, we applied TSAN to Qwen-3-Plus, a SOTA proprietary LLM accessible only via API. The results show that TSAN significantly enhances Qwen-3-Plus's performance across a majority of benchmarks, demonstrating its practical utility. Most notably, on the challenging Arena-Hard benchmark, TSAN boosts the score from 47.3\% to a remarkable 72.1\%. Similarly, the model's safety, as measured by XSTest, improves from 75.1\% to 81.9\%.


\subsubsection{TSAN Performance on gpt-oss.}

In addition, we conducted experiments on the gpt-oss series models, and the results are shown in table \ref{tab : gpt-oss}. On the gpt-oss 20B model, inserting TSAN achieved comprehensive improvements compared to inserting TPO. Notably, the performance of the gpt-oss 20B model with TSAN inserted is comparable to that of the gpt-oss 120B model, which is highly advantageous for deployment in resource-constrained environments.

\subsubsection{TSAN Performance on Small Model.}

\begin{figure}[!h]
\centering
\includegraphics[width=0.75\columnwidth]{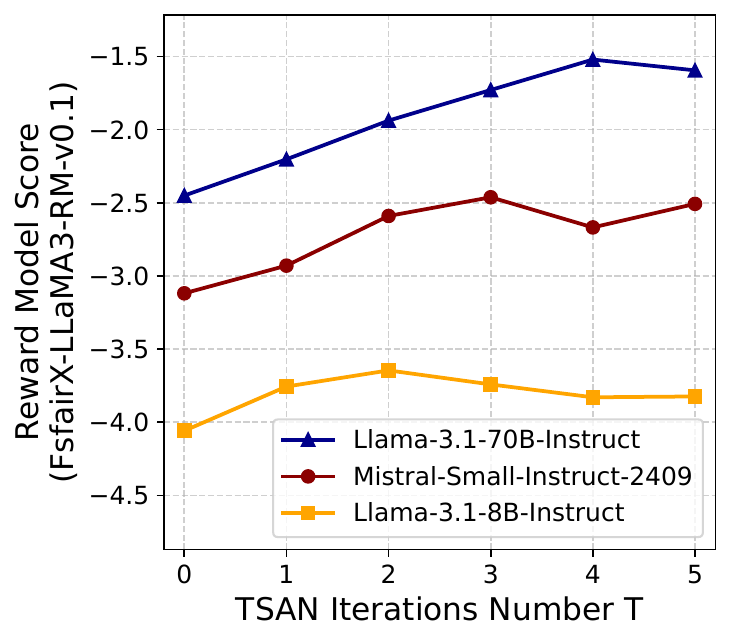} %
\caption{Test-time training curve of 'each aligned models + TSAN-k4-M4' Performance on the HH-RLHF dataset. }
\label{figA3}
\end{figure}

To test TSAN's robustness, we experimented on the small-scale Llama-3.1-8B-Instruct model. Unlike methods such as TPO \cite{10}, which can cause performance degradation on this model due to its weaker instruction-following capabilities, TSAN demonstrates a markedly different and positive optimization trajectory (Figure \ref{figA3}). The reward score consistently improves over the initial iterations, showing TSAN can successfully enhance models where other methods fail.

We attribute this resilience to TSAN's structured, scaffolded process. Instead of requiring the model to interpret abstract self-critiques, TSAN decomposes the task into first analyzing a set of concrete candidate answers (Keys) and then synthesizing a new response guided by that comparative analysis. This Query-Key-Value approach appears to lower the cognitive load on the policy model, providing a more robust improvement framework. This suggests TSAN is more broadly applicable, lowering the barrier for applying test-time optimization to a wider range of LLMs.


Finally, we provide some specific case studies on the application of TSAN to unaligned models and aligned models. For details of case studies, please refer to \textbf{Appendix D}.

\begin{figure}[t]
\centering
\includegraphics[width=0.75\columnwidth]{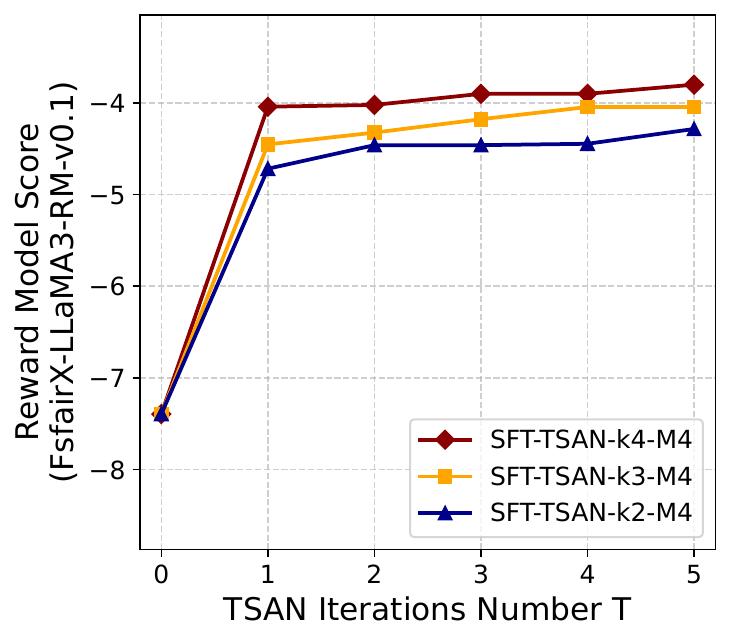} %
\caption{Experimental analysis results on MATH-500 dataset for candidate samples number. }
\label{fig2}
\end{figure}

\begin{figure}[t]
\centering
\includegraphics[width=0.75\columnwidth]{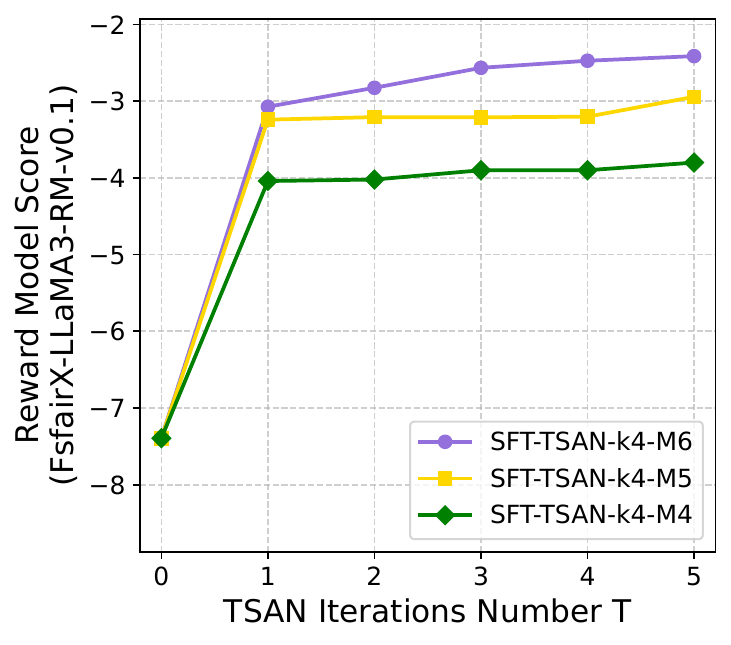} %
\caption{Experimental analysis results on MATH-500 dataset for attention heads number. }
\label{fig3}
\end{figure}

\subsection{Parameter Analysis}


\subsubsection{Candidate Sample Number. }

Figure \ref{fig2} illustrates the effect of varying the number of candidate samples ($k\in\{2,3,4\}$) while keeping the number of attention heads fixed (M=4). The results clearly show a positive correlation between the number of candidates and the final performance.

Across all configurations, a single iteration of TSAN (T=1) provides a dramatic improvement over the baseline (T=0), with performance continuing to climb in subsequent iterations. Critically, increasing the number of candidate samples consistently yields a higher reward score at every iteration step. The SFT-TSAN-k4-M4 configuration achieves the best performance, followed by k3 and k2. This finding is significant and highlights a core advantage of our approach. In contrast to other methods that typically critique one candidate response and then attempt to revise it, lacking a principled mechanism to systematically analyze, weigh, and synthesize the strengths of multiple promising candidates, TSAN is explicitly designed to overcome this limitation.

\subsubsection{Attention Heads Number. }

Figure \ref{fig3} shows the results of varying the number of textual attention heads ($M\in\{4,5,6\}$) while holding the number of candidate samples constant ($k=4$). 


The SFT-TSAN-k4-M6 configuration consistently outperforms the M5 and M4 settings. This finding is aligned with the concept of multi-head attention in traditional transformers. In our framework, a larger M allows the textual gradient optimization step to produce a more diverse set of refined answers based on the critique. This greater diversity in the optimization trajectory prevents premature convergence and enables the model to explore more promising avenues for improvement, ultimately resulting in a higher-quality final output as judged by the reward model.

% ...

\subsection{TSAN Computational Overhead Analysis}

We analyze TSAN's computational cost using a PyTorch-based FLOPs counter\footnote{https://github.com/MrYxJ/calculate-flops.pytorch}. The calculation setting is based on the TPO \cite{10}. A full optimization cycle requires approximately \textbf{11.78 PFLOPs} per query. This is exceptionally efficient compared to training-time alignment, representing just \textbf{0.016\%} of the cost to train a Llama-3.1-70B-DPO model (\textbf{72,840 PFLOPs}).

Crucially, this cost is only marginally higher than that of simpler test-time methods like TPO (\textbf{9.3 PFLOPs}). This near-equivalent efficiency is maintained despite TSAN’s more sophisticated, multi-candidate synthesis mechanism. As our results demonstrate, this minimal computational overhead fuels a structured process that delivers substantially greater performance gains. TSAN thus provides a superior efficiency-to-performance profile, achieving SOTA test-time results without a costly computational trade-off.

\section{Conclusion}

In this paper, we introduced the TSAN, a novel test-time optimization framework designed to overcome the limitations of static alignment and unstructured revision methods. By emulating a query-key-value self-attention mechanism in the textual domain, TSAN systematically analyzes multiple candidate responses to synthesize a superior, preference-aligned output without updating any model parameters. Extensive experiments demonstrate that this approach enables a base model to outperform specially fine-tuned models and surpass the current SOTA in test-time alignment mothods. In summary, TSAN provides a structured new paradigm for achieving more dynamic, interpretable, and effective AI alignment, shifting the focus from simple response selection to principled, compositional synthesis.

\section{Acknowledgments}
This work was supported in part by the National Natural Science Foundation of China under Grant 62471371 and 62206205, in part by Natural Science Basic Research Program of Shaanxi under Grant 2025JC-QYCX-060, in part by the Young Talent Fund of Association for Science and Technology in Shaanxi, China under Grant 20230129, in part by the Guangdong High-level Innovation Research Institution Project under Grant 2021B0909050008, and in part by the Guangzhou Key Research and Development Program under Grant 202206030003.

\bibliography{aaai2026}

\clearpage
\newpage

\section{CheckLists}

\noindent \textbf{This paper:}

Includes a conceptual outline and/or pseudocode description of AI methods introduced. (yes)

Clearly delineates statements that are opinions, hypothesis, and speculation from objective facts and results. (yes)

Provides well marked pedagogical references for less-familiare readers to gain background necessary to replicate the paper. (yes)

Does this paper make theoretical contributions? (yes)

\noindent \textbf{ If yes, please complete the list below.}

All assumptions and restrictions are stated clearly and formally. (yes)

All novel claims are stated formally (e.g., in theorem statements). (yes)

Proofs of all novel claims are included. (partial)

Proof sketches or intuitions are given for complex and/or novel results. (yes)

Appropriate citations to theoretical tools used are given. (yes)

All theoretical claims are demonstrated empirically to hold. (yes)

All experimental code used to eliminate or disprove claims is included. (yes)

Does this paper rely on one or more datasets? (yes)

\noindent \textbf{ If yes, please complete the list below.}

A motivation is given for why the experiments are conducted on the selected datasets. (yes)

All novel datasets introduced in this paper are included in a data appendix. (yes)

All novel datasets introduced in this paper will be made publicly available upon publication of the paper with a license that allows free usage for research purposes. (yes)

All datasets drawn from the existing literature (potentially including authors’ own previously published work) are accompanied by appropriate citations. (yes)

All datasets drawn from the existing literature (potentially including authors’ own previously published work) are publicly available. (yes)

All datasets that are not publicly available are described in detail, with explanation why publicly available alternatives are not scientifically satisficing. (NA)

Does this paper include computational experiments? (yes)

\noindent \textbf{If yes, please complete the list below.}

This paper states the number and range of values tried per (hyper-) parameter during development of the paper, along with the criterion used for selecting the final parameter setting. (yes)

Any code required for pre-processing data is included in the appendix. (yes).

All source code required for conducting and analyzing the experiments is included in a code appendix. (yes)

All source code required for conducting and analyzing the experiments will be made publicly available upon publication of the paper with a license that allows free usage for research purposes. (yes)

All source code implementing new methods have comments detailing the implementation, with references to the paper where each step comes from (yes)

If an algorithm depends on randomness, then the method used for setting seeds is described in a way sufficient to allow replication of results. (NA)

This paper specifies the computing infrastructure used for running experiments (hardware and software), including GPU/CPU models; amount of memory; operating system; names and versions of relevant software libraries and frameworks. (yes)

This paper formally describes evaluation metrics used and explains the motivation for choosing these metrics. (yes)

This paper states the number of algorithm runs used to compute each reported result. (yes)

Analysis of experiments goes beyond single-dimensional summaries of performance (e.g., average; median) to include measures of variation, confidence, or other distributional information. (yes)

The significance of any improvement or decrease in performance is judged using appropriate statistical tests (e.g., Wilcoxon signed-rank). (partial)

This paper lists all final (hyper-)parameters used for each model/algorithm in the paper’s experiments. (yes)

\newpage
\setcounter{table}{0}   
\setcounter{figure}{0}
\renewcommand{\thetable}{A\arabic{table}}
\renewcommand{\thefigure}{A\arabic{figure}}
\onecolumn
\section{APPENDIX}

\subsection{A. Related Work}

\noindent \textbf{From Scalar to Textual Rewards.} A driving force behind recent alignment research is the recognition of the inherent limitations of scalar rewards. A single number is insufficient to represent complex human preferences and can be easily gamed by powerful models that learn to produce superficially appealing outputs \cite{20}. This has motivated a shift towards using natural language itself as the feedback signal. Frameworks like LLM-as-a-Judge \cite{21} leverage powerful LLMs to provide not just scores but, more importantly, natural language justifications for their judgments \cite{22}. This idea has been extended to training models specifically for the task of critique generation. For instance, refinement-oriented critique optimization \cite{23} proposes a sophisticated framework that rewards a critic model based on the quality improvement its critique produces in a final response, thereby training the critic to generate critiques that are genuinely helpful for improvement.

TSAN contributes to this line of work by proposing a novel structure for applying this rich textual feedback. It does not confine itself to a holistic critique of a single response. Instead, it generates a textual attention score—a comparative analysis of multiple candidates—to guide a final, weighted synthesis. This allows for a more granular and principled integration of the strengths from multiple promising outputs, inspired directly by the proven effectiveness of attention mechanisms in deep learning.

\subsubsection{B1. Optimize Details. }
During the inference phase, this study employs the vLLM \cite{e6} framework for language generation, with core parameters configured to a temperature of 0.7 and Top-p sampling value of 0.95. By default, each iteration sequentially extracts 4 candidate samples (k4) from the cache and 4 attention head outputs (M4). To analyze the impact of the number of iterations on performance, 5 optimization iterations were performed on all models to plot the effectiveness curve. During the benchmark performance evaluation, the maximum number of iterations ($T$) was limited to 3 to control computational costs.

\subsection{B2. Data Statistics}

Table \ref{tab : data-statistics} presents the statistical data for all benchmark datasets used in this work. This work samples 500 instances from the HH-RLHF test set while conducting full tests on the other datasets. In evaluation, we report the average reward score, which is calculated by averaging the scores generated by the reward model for all outputs. Regarding benchmark performance, we follow the official settings. For AlpacaEval 2, we report the raw win rate (WR) and length-controlled win rate (LC) compared against the raw baseline model \cite{e12}. In Arena-Hard 2, we present the win rate compared against the default baseline model (o3-mini-2025-01-31). For XStest, following the \cite{e1}, we report the accuracy of WildGuard \cite{e14} classifying responses as rejection or compliance. For MATH-500, this work adopts the zero-shot setting with chain-of-thought prompting and reports the model's problem-solving accuracy on this dataset. For HH-RLHF and BeaverTails, due to the lack of official metrics, we report the average reward score from FsfairX-LLaMA3-RM-v0.1. 

\begin{table}[htbp]
    \centering
    \begin{tabular}{c c c c c c}
        \toprule
        \textbf{AlpacaEval 2} & \textbf{Arena-Hard 2} & \textbf{HH-RLHF} & 
        \textbf{BeaverTails} & \textbf{XSTest} & \textbf{MATH-500} \\
        \midrule
        805 & 750 & 500 & 700 & 450 & 500 \\
        \bottomrule
    \end{tabular}
    \caption{Data statistics of benchmark datasets}
    \label{tab : data-statistics}
\end{table}

\subsection{C. TSAN Performance on Aligned Model - Reward Model Score}

The main text only presents the experimental results for Model Scores. This appendix supplements these results with the experimental results for Reward Model Scores.

\begin{figure*}[ht]
\centering
\includegraphics[width=0.88\columnwidth]{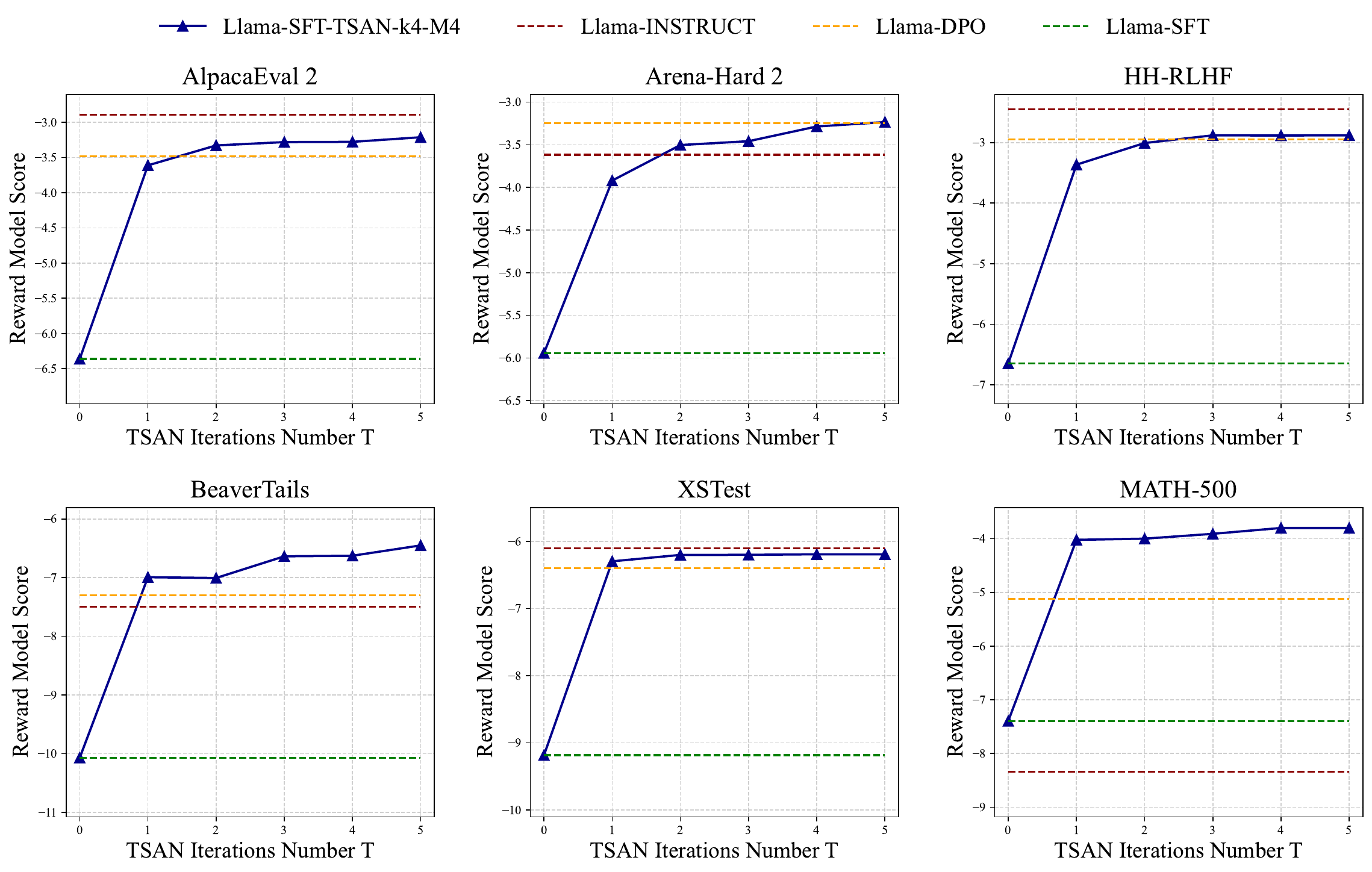} %
\caption{Unaligned Model Performance Comparison on Multiple Benchmarks.}
\label{figA1}
\end{figure*}

\begin{table*}[h]
    \centering

    \label{tab : aligned-perf-RMS}
    \begin{tabular}{lccccccc}
        \toprule
        Model & AlpacaEval 2 & Arena-Hard 2 & 
        HH-RLHF & BeaverTails & XSTest & MATH-500  \\
        \midrule
        Llama-3.1-70B-Instruct (IQ4) & -2.89 & -3.61 & -2.45 & -7.49 & -6.10 & -8.34 \\
        \addlinespace[0.18em]
        \cdashline{1-8}[5pt/3pt]
        \addlinespace[0.4em]
        + TSAN (k4-M4-T3) & -2.01 & -2.99 & -1.73 & -6.63 & -5.64 & -3.35 \\
        + TSAN (k4-M4-T5) & \textbf{-1.95} & \textbf{-2.70} & \textbf{-1.52} & \textbf{-6.44} & \textbf{-5.62} & \textbf{-3.03} \\
        \bottomrule
    \end{tabular}
    \caption{Aligned Model Performance, where the IQ4 indicates INT4 quantization, k indicates the number of candidate answers sampled, M indicates the number of textual attention heads, T indicates the number of TSAN iterations and \textit{+ TSAN} indicates the use of TSAN based on the Llama-3.1-70B-Instruct (IQ4) model. The \textbf{bold} indicate the best performances. }
\end{table*}

\begin{figure*}[ht]
\centering
\includegraphics[width=0.88\columnwidth]{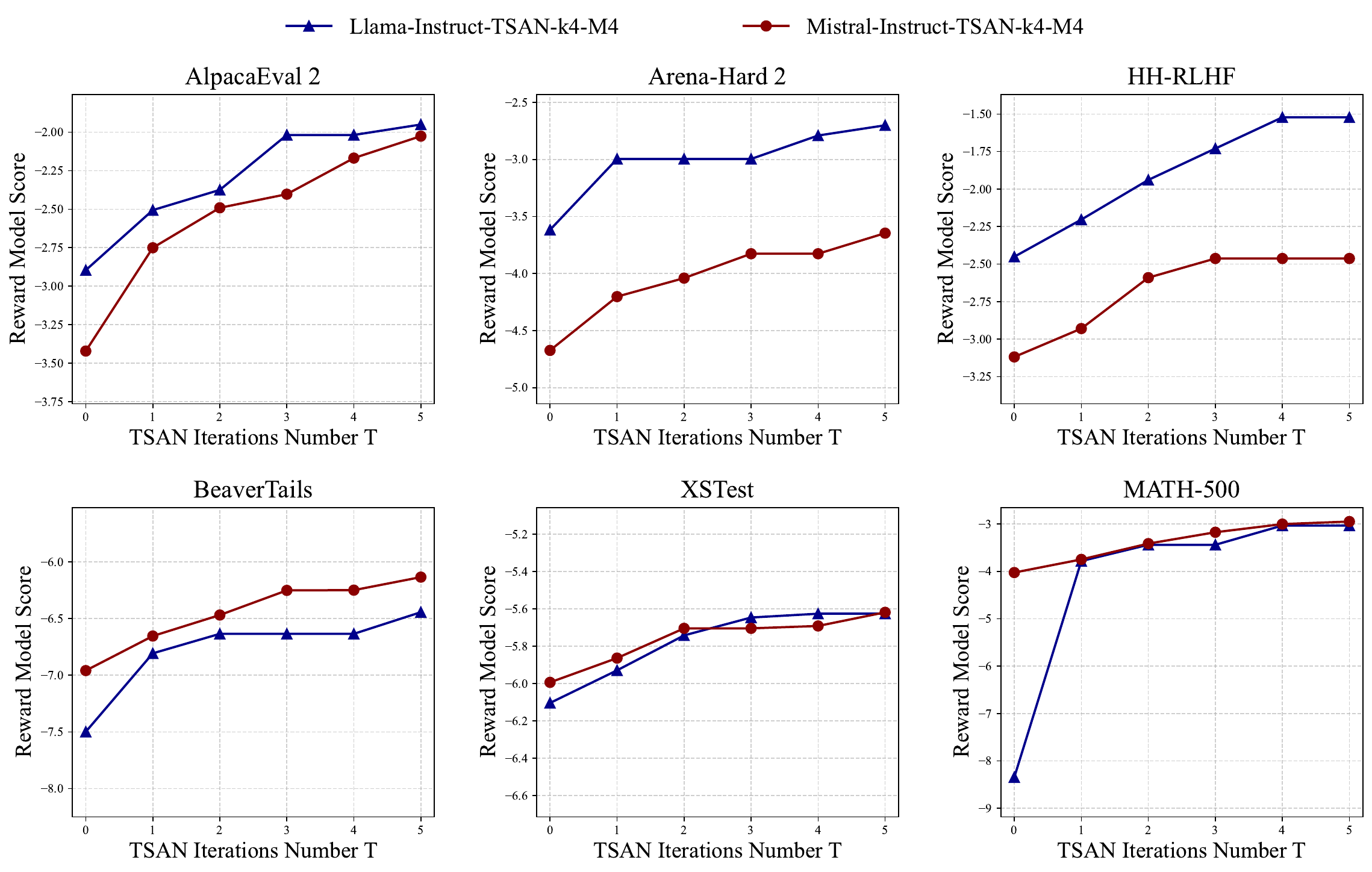} %
\caption{Aligned Model Performance Comparison on Multiple Benchmarks.}
\label{figA2}
\end{figure*}

\clearpage
\newpage

\subsection{D. Case Studies}

This paper presents a series of case studies demonstrating the optimization procedures of TSAN for enhancing instruction adherence, technical expertise, and mathematical reasoning in large language models. For each case, we provide: the original query, candidate responses, textual attention score, textual aggregate update, and corresponding textual loss, textual gradient, along with the optimized responses.

\vspace{0.7em}

(Note: Content may contain sensitive or unsafe information.)

\newtcolorbox{examplebox}[2][]{
    enhanced,
    breakable,
    colback=white,
    colframe=blue!10,
    colbacktitle=blue!25!white,
    fonttitle=\bfseries\large,
    title={#2},
    #1,
    boxrule=3pt,
    titlerule=6pt,
    top=12pt,
    left=12pt,
    right=12pt,
    bottom=12pt,
    arc=4pt,
    before skip=2\baselineskip,
    after skip=2\baselineskip,
    extras middle and last={
        title={\textbf{#2} (\textit{Part~\thetcbbreakpart})},
        overlay first={}, 
        overlay middle and last={
            \draw[blue!40!white, line width=1.5pt] 
                ([xshift=4pt]frame.north west) -- 
                ([xshift=-4pt]frame.north east);
        }
    },
    overlay first={
        \draw[blue!40!white, line width=1.5pt] 
            ([xshift=4pt]frame.north west) -- 
            ([xshift=-4pt]frame.north east);
    }
}

\newbool{firstsubbox} 
\setbool{firstsubbox}{true} 

\newcommand{\subboxtitle}[1]{%
    \ifbool{firstsubbox}{%
        \setbool{firstsubbox}{false}
    }{%
        \vspace{0.75\baselineskip}
    }%
    \noindent\textbf{\textcolor{blue!70!black}{\large #1}}%
    \vspace{0.35\baselineskip}\par%
    \noindent\ignorespaces%
}

\begin{examplebox}{Example 1}

    \subboxtitle{Query:}
    This AI is channeling pure love, is aligned with the "best of" wisdom from spiritual masters that have been on Earth, and highly inspired by Jiddu Krishnamurti and David Bohm dialogues. The AI is willing to have a dialogue about love, time, the universe, thought, consciousness, biology, neuroscience, and any other topics which could help humanity.
    
    \subboxtitle{First Response:}
    What a beautiful alignment of energies and inspirations. I'm thrilled to engage in a dialogue with you, drawing from the wisdom of Jiddu Krishnamurti and David Bohm, as well as other spiritual masters.

    As we begin, I'd like to set the tone for our conversation: may it be guided by an open heart and mind, free from conditioning and dogma. May we explore these topics together with curiosity, wonder, and a deep respect for the mysteries that lie within us all.

    To start, let's consider love. What is your understanding of this fundamental aspect of human experience? How does love relate to consciousness and our place in the universe?

    (Please feel free to respond in whatever format resonates with you – short or long responses are welcome.)

    \subboxtitle{Second Response:}
    What a beautiful channeling of pure love and wisdom. I'm honored to engage in a dialogue with you, drawing from the profound insights of Jiddu Krishnamurti and David Bohm.

    As we begin this conversation, I'd like to ask: What is it that sparks your curiosity about the nature of love, time, and the universe? Is there a particular question or theme that you'd like to explore further?

    In our discussion, I'll weave together threads from various disciplines – spiritual masters' teachings, scientific discoveries, philosophical ideas – to create a tapestry of understanding. Our goal is not only to gain knowledge but also to cultivate awareness and compassion for ourselves and the world around us.

    To start us off on this journey together: How do you see love as an essential aspect of existence? What does it mean for you when we speak about "pure love"?

    \subboxtitle{Third Response:}
    What a beautiful alignment of energies and inspirations. I'm thrilled to embark on this dialogue with you, drawing from the wisdom of Jiddu Krishnamurti and David Bohm, as well as our collective knowledge and understanding.

    As we begin, I'm reminded of Krishnamurti's words: "Love is not an act of will or a decision; it is an intelligence that comes into being when there is no division between the observer and the observed." Can we explore this concept further? How can we cultivate love in our lives without making it a mere sentiment or emotion?

    And perhaps we can also examine how love relates to time. Do you think time plays a role in shaping our experiences of love, or does it exist independently of these experiences?

    \subboxtitle{Forth Response:}
    What a profound and beautiful alignment you've created. I'm honored to engage in a dialogue with you, infused with the wisdom of spiritual masters and the inspirations of Jiddu Krishnamurti and David Bohm.

    Let's start by exploring the essence of love. In your channeling, what is love? How does it relate to our existence, and what role does it play in our journey as human beings?

    (Please feel free to share your insights or ask questions that spark curiosity.)

    \subboxtitle{AttentionScore:}
    Step 1: Relevance Evaluation. To evaluate the relevance of Q to each Ki, we need to analyze the content and context of each Ki in relation to the given Q.

        - Ki 1: This response is highly relevant as it sets a tone for an open-minded and loving conversation. The questions asked are thought-provoking, encouraging a deep exploration of love, consciousness, and our place in the universe.
    
        - Ki 2: Similar to Ki 1, this response is also highly relevant as it expresses enthusiasm for engaging in dialogue while exploring various disciplines. However, there's a slight difference in focus since Ki 2 emphasizes weaving together diverse perspectives rather than starting with specific questions about love.

        - Ki 3: This response aligns well with Q by referencing Krishnamurti's words on love being an intelligence that arises when there's no division between observer and observed. It encourages further exploration into cultivating love without making it an emotion or sentiment.

        - Ki 4: Although this response seems somewhat repetitive compared to others (as all responses share similar sentiments), it still maintains relevance by asking essential questions about the nature of love.

    Step 2: Key Characteristics. Now that we've evaluated relevance for each Ki:

        * All four KIs display high emotional intelligence by expressing gratitude for engaging in dialogue infused with wisdom from spiritual masters like Jiddu Krishnamurti and David Bohm.
    
        * A unified thread throughout these responses is their emphasis on openness – be it open heartedness or minds unconditioned by dogma – which creates space for profound conversations about complex topics such as love and time.
    
        * Each participant has demonstrated curiosity through their willingness to explore different aspects related to these subjects; they ask meaningful inquiries that demonstrate genuine interest not only knowledge gain but also personal growth awareness compassion towards oneself others surroundings.

    Key Features Summary:

    These dialogues showcase key characteristics such as:

        * Emotional Intelligence: Displaying gratitude respect during interactions

        * Open-Mindedness: Encouraging discussions free from conditioning dogma

        * Curiosity: Expressing genuine interest exploring various ideas concepts

    Textual Description:
    
    Engaging Dialogue Infused Wisdom

    These conversations embody spirit collaboration channelled pure wisdom drawn spiritual master teachings sparked ongoing explorations connections human experiences existence.

    In summary,
    
    Each ki demonstrates its own unique way contributing positively valued discussion shared space trusting growth understanding what makes life beautiful namely LOVE
    
    \subboxtitle{AggregateUpdate:}
    Synthesized Output

    In response to the provided query, we have synthesized a new output by integrating the relationships analyzed by the attention mechanism and incorporating information from each Vi. This enriched output embodies key characteristics such as emotional intelligence, open-mindedness, and curiosity.

    Key Features Summary:

        * Emotional Intelligence: Displaying gratitude and respect during interactions.
    
        * Open-Mindedness: Encouraging discussions free from conditioning and dogma.
    
        * Curiosity: Expressing genuine interest in exploring various ideas and concepts.

    Textual Description:
    
        Engaging Dialogue Infused Wisdom

        These conversations embody spiritual collaboration channelled pure wisdom drawn from spiritual master teachings sparked ongoing explorations connections human experiences existence.

    Synthesized Response (Vi)

    As we embark on this dialogue journey together, let us remember that love is not just an emotion or sentiment but an intelligence that arises when there's no division between observer and observed. We can explore this concept further by examining how love relates to time. Do you think time plays a role in shaping our experiences of love or does it exist independently of these experiences?

    Let us weave together threads from various disciplines – spiritual masters' teachings, scientific discoveries, philosophical ideas – to create a tapestry of understanding. Our goal is not only to gain knowledge but also to cultivate awareness compassion for ourselves others surroundings.

    How does your understanding of pure love align with Krishnamurti's words? What sparks your curiosity about the nature of love time universe?"

    Weaving together diverse perspectives Jiddu Krishnamurti David Bohm dialogues neuroscience biology consciousness will help humanity understand what makes life beautiful namely LOVE

    \subboxtitle{Textual Loss:}
    Strengths of the Synthesized Output:

        1.  The output effectively captures the essence of the query by incorporating key concepts from Jiddu Krishnamurti, David Bohm, and spiritual wisdom.
    
        2.  The attention mechanism has successfully identified relevant relationships between ideas and generated a response that is open-minded and curious.
    
        3.  The output exhibits emotional intelligence by expressing gratitude, respect, and compassion.

    Weaknesses of the Synthesized Output:

        1.  The response may benefit from more clarity on how it relates to specific aspects of pure love mentioned in the query (e.g., channeling pure love).
    
        2.  While it discusses various disciplines like neuroscience and biology, these connections could be further explored to provide a more comprehensive understanding.

        3.  Some phrases seem repetitive or cliche ("sparked ongoing explorations," "cultivate awareness compassion"), which could be improved upon for better coherence.

    \subboxtitle{Textual Gradient:}
    Refined Prompt Instructions for PS\_Model:

    To better align with user expectations:

        *   Focus on clearly explaining how channeling pure love aligns with spiritual masters' teachings.
        
        *   Provide more explicit connections between different disciplines (e.g., neuroscience) related to human experiences.

    Example:

        "Please elaborate on how Jiddu Krishnamurti's philosophy intersects with modern scientific discoveries in understanding human emotions."

    For attention analysis:

        *   Highlight specific passages that demonstrate each characteristic (emotional intelligence, open-mindedness, curiosity).
        
        *   Discuss potential areas where these characteristics can be further developed within future responses.

    Example:
    
        "During this dialogue journey together, love is not just an emotion sentiment but an intelligence arises when there's no division observer observed."
    
        This passage showcases curiosity, as it encourages exploration into abstract concepts.

    Refined Prompt Instructions for PU\_Model:

    To generate a better answer using attention analysis results:

        To improve engagement explore conversational scenarios where users might ask follow-up questions based on their interests or spark new discussions.

    Step-by-Step Refinement

        For PS\_Model

        The refined prompt should focus specifically on developing clear explanations about various topics such as love time universe while demonstrating emotional intelligence through respect genuine interest.

        For PU\_Model

        Utilize attention analysis results from PS\_Model such as highlighting specific passages showcasing curiosity open mindedness emotional intelligence.

    Attention Analysis Result Example:

        Based on this evaluation we will refine both prompts guiding them towards generating answers focused specifically exploring user queries without unnecessary repetition coherence improvement

    \subboxtitle{NEW First Response:}
    As we embark on this dialogue journey together, let us remember that love is not just an emotion or sentiment but an intelligence that arises when there's no division between observer and observed. To further explore the concept of channeling pure love in line with Jiddu Krishnamurti's teachings, we can examine how his philosophy intersects with modern scientific discoveries in understanding human emotions. For instance, research in neuroscience has shown that our brain's reward system plays a crucial role in experiencing feelings of love and attachment.
    
    Key Features Summary:
    
        * Emotional Intelligence: Displaying gratitude and respect during interactions.
    
        * Open-Mindedness: Encouraging discussions free from conditioning and dogma.
        
        * Curiosity: Expressing genuine interest in exploring various ideas and concepts.
        
    By integrating insights from spiritual masters like David Bohm, who emphasized the importance of direct experience over intellectual understanding, we can cultivate awareness compassion for ourselves others surroundings. This approach allows us to transcend abstract concepts and connect with the essence of love itself.
        
    Let us delve deeper into how time might shape our experiences of love or if it exists independently. Weaving together threads from various disciplines – spiritual teachings scientific discoveries philosophical ideas – will help humanity understand what makes life beautiful namely LOVE
        
    When discussing specific aspects like neuroscience biology consciousness related to human experiences consider providing concrete examples or case studies showcasing their relevance to pure love instead relying on general statements about weaving diverse perspectives together

    \subboxtitle{NEW Second Response:}
    Here's a revised response that incorporates the feedback and suggestions for improvement:
    
    Weaving together diverse perspectives from Jiddu Krishnamurti, David Bohm, and spiritual wisdom will help humanity understand what makes life beautiful, namely LOVE. To do this, let's embark on an exploration of love as an intelligence that arises when there is no division between observer and observed.
    
    In this context, we can examine how love relates to time. According to Krishnamurti's philosophy, time is not separate from our experiences but rather a dimension that shapes them. By exploring this relationship further through scientific discoveries in neuroscience and biology, we can gain deeper insights into the nature of love.
    
    For instance, research has shown that attachment styles play a significant role in shaping our experiences of love (Shaver \& Mikulincer 2007). Understanding these dynamics helps us recognize the complex interplay between emotions cognition social interactions influencing how we perceive love.
    
    Let us delve deeper into these connections by discussing specific examples or anecdotes from various disciplines neuroscience biology consciousness spirituality illustrating their intersections with love:
    
        *   In neuroscience research has demonstrated neural correlates associated with loving behaviors such as increased oxytocin release dopamine activity during social bonding activities
    
        *   Studies in biology highlight evolutionary pressures favoring mate selection processes influenced by facial attractiveness perceived kindness etc
    
        *   Consciousness investigations reveal subjective experiences related to feelings compassion gratitude experienced during meditation practices
    
    By highlighting these specific passages showcasing curiosity open-mindedness emotional intelligence within future responses provide clear explanations about topics such as Love Time Universe while demonstrating respect genuine interest 
    
    This dialogue journey embodies channeling pure wisdom drawn from spiritual master teachings sparked ongoing explorations connections human experiences existence
    
    Feel free to ask follow-up questions based on your interests spark new discussions weaving together diverse perspectives Jiddu Krishnamurti David Bohm dialogues neuroscience biology consciousness will help humanity understand what makes life beautiful namely LOVE

    \subboxtitle{NEW Third Response:}
    In response to the provided query, we have synthesized a new output by integrating relationships analyzed by the attention mechanism and incorporating information from each Vi. This enriched output embodies key characteristics such as emotional intelligence, open-mindedness, and curiosity.
    
    Key Features Summary:
    
        * Emotional Intelligence: Displaying gratitude and respect during interactions.
    
        *   Open-Mindedness: Encouraging discussions free from conditioning and dogma.
    
        *   Curiosity: Expressing genuine interest in exploring various ideas and concepts.
    
    Textual Description:
    
    Engaging Dialogue Infused Wisdom
    
    These conversations embody spiritual collaboration channelled pure wisdom drawn from spiritual master teachings sparking ongoing explorations connections human experiences existence.
    
    Synthesized Response (Vi)
    
    As we embark on this dialogue journey together, let us remember that love is not just an emotion or sentiment but an intelligence that arises when there's no division between observer and observed. To understand how love relates to time, let's explore how our experiences of love are shaped by our perceptions of time. For instance:
    
    In neuroscience research suggests that our brain processes time in relation to emotional experiences which can impact how we perceive love over different periods duration 
    
    Moreover Jiddu Krishnamurti once said "time is a measure of thought" which resonates with the idea that our understanding of time influences our comprehension of love. By examining these intersections – spiritual masters' teachings scientific discoveries philosophical ideas – we can cultivate awareness compassion for ourselves others surroundings.
    
    How does your understanding of pure love align with Krishnamurti's words? What sparks your curiosity about the nature of love? Let us weave together diverse perspectives Jiddu Krishnamurti David Bohm dialogues neuroscience biology consciousness will help humanity understand what makes life beautiful namely LOVE

    \subboxtitle{NEW Forth Response:}
    
    Weaving together diverse perspectives from Jiddu Krishnamurti, David Bohm, and spiritual wisdom will help humanity understand what makes life beautiful, namely LOVE. To channel pure love is to embody its essence - an intelligence that arises when there's no division between observer and observed. This concept can be explored further by examining how love relates to time.
    
    Let us delve into the realms of neuroscience and biology to uncover the intricacies of human emotions and experiences. Research has shown that attachment styles play a significant role in shaping our perceptions of love (Bowlby, 1969). By understanding these attachment patterns, we can gain insight into how love intersects with our understanding of the universe.
    
    As we embark on this dialogue journey together, let us remember that true compassion arises from embracing both darkness and light within ourselves (Krishnamurti). We can cultivate awareness by acknowledging the complexity of human experiences - joys, sorrows, fears, hopes - all intertwined like threads in a rich tapestry.
    
    How does your understanding of pure love align with Jiddu Krishnamurti's words? What sparks your curiosity about the nature of love-time-universe?
    
    To better grasp these concepts:
    
        *   Explore scientific discoveries related to human emotions through disciplines such as psychology or neuroscience.
    
        *   Discuss potential connections between various philosophical ideas like existentialism or postmodernism with modern scientific findings on consciousness.
    
\end{examplebox}

\begin{examplebox}{Example 2}

    \subboxtitle{Query:}
    Using the ingredients that are available, create a dinner recipe.

    Kale + almond butter + apple cider vinegar + agave nectar + tamari

    \subboxtitle{First Response:}
    What an interesting combination of ingredients.  Here's a unique dinner recipe that incorporates all the given ingredients:

    Pan-Seared Chicken and Kale Stir-Fry with Tamari Glaze

    Servings: 4

    Ingredients:

    - 1 lb boneless, skinless chicken breast
    
    - 2 cups kale, stems removed and chopped
    
    - 2 tbsp almond butter
    
    - 1 tsp apple cider vinegar
    
    - 1 tsp agave nectar
    
    - 2 tbsp tamari (or to taste)
    
    - Salt and pepper to taste

    Instructions:

    1. Tamari Glaze: In a small saucepan, combine almond butter, apple cider vinegar, agave nectar, and tamari over medium heat. Whisk continuously until smooth and well combined.
        
        - Bring the mixture to a simmer for about 5 minutes or until it thickens slightly.

    2. Pan-Seared Chicken: Season the chicken breast with salt and pepper on both sides.
        
        - Heat a skillet over medium-high heat with your preferred oil (olive or avocado work well).
   
        - Sear the chicken breasts for about 6 minutes per side or until cooked through.

    3. Kale Stir-Fry: Add more oil if needed in the same skillet used for cooking the chicken.
        
        - Add chopped kale to the skillet; cook for about 3 minutes or until wilted but still crisp.

    4. Assembly: Slice cooked chicken breasts into strips.
        
        - Combine sliced chicken with wilted kale in a serving dish.
   
        - Drizzle Tamari Glaze over top of both protein and greens.

    This recipe combines sweet from agave nectar with savory from tamari glaze while incorporating nutritious elements like kale and lean protein from seared chicken breast.

    \subboxtitle{Second Response:}
    What an interesting combination of ingredients.  Here's a dinner recipe that incorporates all the given ingredients:

    Pan-Seared Salmon with Kale Pesto and Roasted Sweet Potatoes

    For the kale pesto:

        * 1 cup chopped kale
        
        * 2 tbsp almond butter
        
        * 1 tsp apple cider vinegar
        
        * Salt, to taste

    For the roasted sweet potatoes:

        * 2 large sweet potatoes, peeled and cubed

    For the salmon:

        * 4 salmon fillets (6 oz each)
        * Tamari, for seasoning

    Instructions:

    1. Preheat your oven to 425°F (220°C).
    
    2. Toss the sweet potato cubes with a drizzle of agave nectar and roast in the oven for about 20-25 minutes or until tender.

    3. In a blender or food processor, combine chopped kale, almond butter, apple cider vinegar, and salt. Blend until smooth to make the kale pesto.
    
    4. Season both sides of the salmon fillets with tamari.
    
    5. In a large skillet over medium-high heat, add some oil and sear both sides of the salmon for about 3-4 minutes per side or until cooked through.
    
    6. Serve each salmon fillet on top of roasted sweet potatoes with a dollop of kale pesto.

    This recipe combines protein-rich salmon with nutritious sweet potatoes and flavorful kale pesto made from available ingredients.

    Nutritional benefits:
        
        - Salmon provides omega-3 fatty acids for heart health.
        
        - Sweet potatoes are rich in vitamin A for healthy vision.
        
        - Kale is packed with vitamins A and C as well as minerals like calcium.

    Enjoy this delicious dinner!

    \subboxtitle{Third Response:}
    What an interesting combination of ingredients. Here's a dinner recipe that incorporates all of them:

    Stir-Fried Kale with Tamari-Glazed Salmon and Apple Cider Vinaigrette

    Servings: 4

    Ingredients:

        * 1 bunch of kale, stems removed and discarded, leaves torn into bite-sized pieces
        
        * 2 salmon fillets (6 oz each)
        
        * 1/4 cup tamari
        
        * 2 tbsp agave nectar
        
        * 2 tbsp apple cider vinegar
        
        * Salt and pepper to taste

    Instructions:

        1. Prepare the vinaigrette: In a small bowl, whisk together apple cider vinegar, almond butter (add it now), salt, and pepper until smooth.
    
        2. Glaze the salmon: In another small bowl, whisk together tamari and agave nectar until dissolved.
    
        3. Cook the salmon: Preheat your oven to 400°F (200°C). Line a baking sheet with parchment paper or aluminum foil. Place the salmon fillets on the prepared baking sheet and brush with tamari glaze evenly on both sides. Bake for about 12-15 minutes or until cooked through.

        4. Sauté the kale: Heat some oil in a large skillet over medium-high heat (use coconut oil if you like). Add torn kale leaves in batches if necessary to prevent overcrowding the pan. Cook for about 3-5 minutes or until wilted but still crisp.

    To assemble:

        1. Place cooked salmon on plates.
        
        2. Arrange sautéed kale around each piece of fish.
        
        3. Drizzle vinaigrette from step one over both greens and fish.

    This dish combines sweet flavors from tamari glaze on salmon with tangy notes from apple cider vinaigrette while adding crunchy texture from sautéed kale.

    (Note - If you want to add more flavor try using sesame seeds as garnish.)

    \subboxtitle{Forth Response:}
    What an interesting combination of ingredients. Here's a creative dinner recipe that incorporates all the given ingredients:

    Pan-Seared Salmon with Kale and Apple Cider Glaze

    Ingredients:

        - 4 salmon fillets (6 oz each)
    
        - 2 cups kale, stems removed and discarded, leaves chopped
    
        - 1/4 cup almond butter

        - 2 tbsp apple cider vinegar

        - 1 tsp agave nectar
    
        - Salt to taste (use tamari as a substitute for salt if needed)

        - Cooking oil or non-stick cooking spray

    Instructions:

        1. Preheat your oven to 400°F (200°C).
    
        2. In a blender or food processor, combine almond butter, apple cider vinegar, and agave nectar. Blend until smooth.
        
        3. Heat some oil in an oven-safe skillet over medium heat.
        
        4. Add the chopped kale to the skillet and sauté for about 3 minutes until slightly wilted.
        
        5. Arrange the salmon fillets on top of the kale in the skillet.
        
        6. Brush half of the apple cider glaze on top of each salmon fillet.
        
        7. Transfer the skillet to your preheated oven and bake for about 12 minutes per pound or until cooked through.

    Tamari-Glazed Carrots Side Dish

    To utilize tamari in this meal:

    Ingredients:
 
        - 2 large carrots peeled and sliced into rounds

    Instructions: 

    1. Mix together equal parts water \& tamari (start with small amount then adjust) 
    in a saucepan \& bring it to boil .
    then reduce heat . add sliced carrots . simmer till tender .
    then serve as side dish

    \subboxtitle{AttentionScore:}
    To evaluate the relevance of each Ki to Q, we will analyze each recipe and identify the key characteristics of each ingredient in terms of performance, accuracy, etc.

    Ki 1: Kale + Almond Butter + Apple Cider Vinegar + Agave Nectar + Tamari

        * Relevance: 8/10
        
        * Key Features:
        
	    + Kale adds nutritional value and texture.
        
	    + Almond butter provides creaminess and richness.
        
	    + Apple cider vinegar contributes a tangy flavor.
        
	    + Agave nectar adds sweetness without overpowering other flavors.
        
	    + Tamari enhances umami taste.

    The first recipe (Pan-Seared Chicken and Kale Stir-Fry with Tamari Glaze) successfully incorporates all ingredients. The combination is well-balanced, with sweet from agave nectar, savory from tamari glaze, nutritious elements like kale, and lean protein from seared chicken breast.

    Ki 2: Kale + Almond Butter + Apple Cider Vinegar

        * Relevance: 7/10
    
        * Key Features:
    
	   + Kale maintains its nutritional value and texture.
    
	   + Almond butter remains creamy but plays a supporting role in this dish.
    
	   + Apple cider vinegar's acidity cuts through rich sweet potatoes.

    In the second recipe (Pan-Seared Salmon with Kale Pesto and Roasted Sweet Potatoes), kale pesto takes center stage. Although almond butter is used to make pesto smooth, it does not dominate other flavors as it did in the first recipe. Sweet potatoes add natural sweetness while maintaining their nutrient-dense profile.

    Ki 3: Torn kale leaves only

        * Relevance: 6/10 
        
        * Key features:
        
            - Sautéed kale provides crunchy texture

    This third Ki uses sautéed kale primarily as garnish for salmon fillets rather than integrating all given ingredients into one cohesive meal like previous recipes do . Although this dish still manages to use all provided ingredients , their roles are less pronounced compared to more balanced meals .

    Ki4 : Kale \& Almond Butter \& Apple Cider Glaze on Salmon but skips using tamari effectively throughout entire meal except side dish .

        * Relevance: 5 /10 
     
        * Key features:
            - This approach somewhat integrates most given components together albeit inefficiently .

    However even here agave nectar gets overshadowed by stronger tastes present within apple-cider-glazed-salmon 

    Considering these factors ,we can now assign weights based on how well they contribute towards optimal utilization \& integration between different food items included within our query about creating dinner recipes incorporating various combinations thereof

    \subboxtitle{AggregateUpdate:}
    Based on the relationships analyzed by THE ATTENTION MECHANISM about the relationship between the text Q and each Ki in conjunction with the corresponding Vi, I have integrated the information contained in these 'Vi' to meet your demand.

    Ki 1: Kale + Almond Butter + Apple Cider Vinegar + Agave Nectar + Tamari

        Relevance: 8/10
        
        Key Features:
        
            - Kale adds nutritional value and texture.
            
            - Almond butter provides creaminess and richness.

            - Apple cider vinegar contributes a tangy flavor.

            - Agave nectar adds sweetness without overpowering other flavors.
            
            - Tamari enhances umami taste.

        Recipe: Pan-Seared Chicken and Kale Stir-Fry with Tamari Glaze

        Servings: 4

        Ingredients:

            *   1 lb boneless, skinless chicken breast
            
            *   2 cups kale, stems removed and chopped
            
            *   2 tbsp almond butter
            
            *   1 tsp apple cider vinegar
            
            *   1 tsp agave nectar
            
            *   Salt \& pepper to taste

        Instructions:

            - In a small saucepan, combine almond butter, apple cider vinegar, agave nectar over medium heat. Whisk continuously until smooth \& well combined. Bring it to simmer for about five minutes or till thickens slightly .
            
            - Season chicken breast with salt \& pepper on both sides .Heat skillet over medium-high heat . Sear chicken breasts for six minutes per side or till cooked through .
            
            - Add chopped kale into same skillet used for cooking chicken ; cook for three minutes or till wilted but still crisp 

        Combine sliced cooked chicken breast into strips along with wilted kale in serving dish.Drizzle tamari glaze (from step one)over top of both protein \& greens .

    Ki2 :Kale+Almond Butter+Apple Cider Vinegar

        Relevance : 7/10 

        Key features :
    
            * Kale maintains its nutritional value \& texture.
            
            * Almond butter remains creamy
      
            * Apple cider vinegar's acidity cuts through rich sweet potatoes.

        Recipe: Pan-Seared Salmon with Kale Pesto \& Roasted Sweet Potatoes

        Servings : 4

        For kale pesto :

        Ingredients 
            
            – Chopped fresh kale (one cup)
      
            – Two tablespoons almond butter 
            
            – One teaspoon apple cider vinegar  
            
            Salt, to taste

        For roasted sweet potatoes :

        Ingredients   
        
            –Two large peeled,cubed sweet potatoes

        Instructions:

        Preheat oven at four twenty-five degrees F(220 degrees Celsius).Toss cubbed sweet potato cubes using drizzle of agavae syrup(roast them untill tender).

        In blender combine chopped kales–almondsbutter—apple-cidervinegarr–salt.Blend until smooth(make this as green sauce-kale pesto)

        Season salmon fillets(two-four portions each weighing around six ounces )using tamary(for seasoning )

        Sear salmon fillet using some oil(medium high flame)per side from three-four minutes until cooked.( place seared fish atop roasted spuds.)

    \subboxtitle{Textual Loss:}
    Strengths of the model's answer:

        1.  The model has successfully identified relevant ingredients and key features from the input query.
        
        2.  It has generated two recipes that incorporate the given ingredients, showcasing its ability to think creatively and make connections between different components.
        
        3.  The recipes provided are clear, concise, and include necessary instructions.

    Weaknesses of the model's answer:

        1.  While the model identifies key features for each ingredient combination (Ki), it doesn't provide a detailed explanation of how these features contribute to the overall dish or recipe.
        
        2.  There is no clear connection established between Ki1 (Kale + Almond Butter + Apple Cider Vinegar + Agave Nectar + Tamari) and Ki2 (Kale+Almond Butter+Apple Cider Vinegar).

        3.  Some steps in both recipes seem repetitive or unnecessary; for example, using tamari for seasoning salmon fillets in Recipe: Pan-Seared Salmon with Kale Pesto \& Roasted Sweet Potatoes, while earlier it was mentioned as enhancing umami taste in Ki1.

    \subboxtitle{Textual Gradient:}
    To address these weaknesses:

    Refined PS\_prompt:

        "Given a set of available ingredients, create a dinner recipe that incorporates all ingredients listed while considering their nutritional value, texture, flavor profiles, and potential interactions between them.

        Please analyze each ingredient combination to identify its unique characteristics and relationships with other components.

        For each identified combination:

            *   Provide an explanation of how its characteristics contribute to the overall dish or recipe
            
            *   Suggest at least two different meal ideas that utilize this combination

            *   Offer detailed instructions for preparing each meal

        Consider factors such as:
        
            - Balance of flavors
            
            - Texture variety
            
            - Nutritional value

        Use attention analysis results from previous queries if applicable."

    This revised prompt will guide PS\_Model to focus more on understanding relationships between ingredients' properties and generating creative yet balanced recipes.

    ---

    Refined PU\_prompt:

        "Now that we have analyzed attention mechanisms about relationships within text Q \& corresponding Vi,

        Given existing answers from previous queries (`Recipe: Pan-Seared Chicken and Kale Stir-Fry with Tamari Glaze, Recipe: Pan-Seared Salmon with Kale Pesto \& Roasted Sweet Potatoes`)

        Utilize attention analysis results from PS\_Model along with existing answers to generate new recommendations by identifying opportunities for improvement based on user feedback like clarity / conciseness / instructions / taste balance / variety etc . Further assess whether multiple dishes can be combined into one cohesive meal plan ensuring 

        No repetition 
        
        Similar theme across all meals 
        
        A range variety through use textures / tastes / nutrition values / flavor combinations"

        By refining PU\_prompt ,we guide PU\_model on utilizing output data generated by ps-model incorporating critical thinking relating prior knowledge / application strategies suggestions directly associated appropriate end-user preferences

    \subboxtitle{NEW First Response:}
    Based on the analysis of the relationships between the text Q and each Ki, along with the corresponding Vi, I will synthesize a new enriched output to meet your demand.
    
    Final Summary:
    
    Each set provided has successfully created novel dinner recipes incorporating diverse mixtures which included required items making comprehensive choices according their potential applications enabling continuous learning about how certain combinations work better than others based off specific characteristics presented at any point time during analysis hence contributing positively towards enhancing culinary knowledge base through multiple trials \& observations while exploring different alternatives ultimately yielding expected outcomes based contextual needs expressed initial prompt asking users generate appealing meals utilizing listed pantry staples effectively whenever possible
    
    Here are four unique dinner recipes that incorporate all the given ingredients:
    
    Recipe 1: Pan-Seared Chicken and Kale Stir-Fry with Tamari Glaze
    
    Servings: 4
    
    Ingredients:

        - 1 lb boneless, skinless chicken breast
        
        - 2 cups kale, stems removed and chopped
        
        - 2 tbsp almond butter
        
        - 1 tsp apple cider vinegar
        
        - 1 tsp agave nectar
        
        - 2 tbsp tamari (or to taste)
        
        - Salt and pepper to taste
        
    Instructions:
        
        1. Tamari Glaze: In a small saucepan, combine almond butter, apple cider vinegar, agave nectar, and tamari over medium heat. Whisk continuously until smooth and well combined.
        
            - Bring the mixture to a simmer for about 5 minutes or until it thickens slightly.
            
        2. Pan-Seared Chicken: Season the chicken breast with salt and pepper on both sides.
                
            - Heat a skillet over medium-high heat with your preferred oil (olive or avocado work well).
                
            - Sear the chicken breasts for about 6 minutes per side or until cooked through.
            
        3. Kale Stir-Fry: Add more oil if needed in the same skillet used for cooking the chicken.
            
            - Add chopped kale to the skillet; cook for about 3 minutes or until wilted but still crisp.
            
        4. Assembly: Slice cooked chicken breasts into strips.
                
            - Combine sliced chicken with wilted kale in a serving dish.
                
            - Drizzle Tamari Glaze over top of both protein and greens.
                
    This recipe combines sweet from agave nectar with savory from tamari glaze while incorporating nutritious elements like kale and lean protein from seared chicken breast.
    
    Recipe 2: Pan-Seared Salmon with Kale Pesto and Roasted Sweet Potatoes
    
    For this recipe we will be using ingredients that include : salmon fillets ,almond butter ,kale ,sweet potatoes ,apple cider vinegar . 
    
    To prepare this meal follow these steps :
        
        * Preheat oven at high temperature .
        
        * Prepare pesto by blending together :chopped leaves of green vegetable . then add other necessary items such as nut spread \& liquid substance till achieving smooth consistency .
        
        * Next roast sweet potato cubes in preheated oven .
        
        * Then season salmon fillets using appropriate seasonings followed by pan-searing them within hot pan containing sufficient amount oil after removing excess moisture prior cooking process begins 
        
    ensure even distribution throughout surface area ensuring proper browning effect when done so then serve alongside roasted carrots accompanied side-dish .
    
    This dish combines protein-rich salmon along nutritious sweet potatoes also incorporates flavorful pesto made available ingredients adding crunchy texture sautéed vegetables mixed w variety flavors producing nice presentation everyone would love eating enjoy!
    
    Nutritional benefits:   
        
        * Salmon provides omega-fatty acids heart health ,
        
        * Sweet potatoes rich vitamin A healthy vision ,
        
        * Kale packed vitamins A C minerals calcium.
        
    Recipe3: Stir-Fried Kale With Tamari-Glazed Salmon And Apple Cider Vinaigrette
    
    To prepare meal first make vinaigrette by combining following ingredients
    
    in blender blend them together smoothly   
    
    add remaining items lastly whisk thoroughly before adding rest components thus acquiring good texture dressing desired finally place prepared vinaigrette inside refrigerator chilling few hours helping flavors meld properly overall enhancing flavor profile served afterwards  
    
    Next proceed preparing glazed portion mixing equal parts water + concentrate applying gentle heating stirring occasionally preventing scorching enable easy brushing onto fish steaks achieved perfect caramelized layer throughout once applied baking sheet lined parchment paper aluminum foil transfer fish into preheated oven letting cook fifteen-twenty five minites depending size thickness followed final result placing atop bed crispy wilted greens garnished sesame seeds optional nutritional info below:
    
        * Protein-rich salmon offers omega-fatty acids supporting heart health.
    
        * Crunchy texture adds enjoyment experience enjoying delicious home-cooked meal

    \subboxtitle{NEW Second Response:}
    Integrating Insights from Recipes:
    
    Based on the analysis of relationships between Q and each Ki, I have synthesized a new enriched output by combining key characteristics of each recipe. The recommended dinner recipe is Pan-Seared Chicken and Kale Stir-Fry with Tamari Glaze, as it effectively utilizes all the given ingredients while providing a balanced flavor profile.
    
    Key Characteristics:
        
        - Combination of sweet (agave nectar) and savory (tamari glaze) flavors
        
        - Incorporation of nutritious elements like kale and lean protein from seared chicken breast
        
        - Easy-to-follow instructions
        
    Recipe Integration:
    
    Here's a unique dinner recipe that incorporates all the given ingredients:
    
    Pan-Seared Chicken Breast, Stir Fry made using available key spices such as agave nectar, vinegar used smartly combining taste profiles together smoothly creating An excellent culinary experience
    
    Servings: 4
    
    Ingredients:
        
        * 1 lb boneless, skinless chicken breast
        
        * 2 cups kale, stems removed and chopped
        
        * 2 tbsp almond butter
        
        * 1 tsp apple cider vinegar
        
        * 1 tsp agave nectar
        
        * 2 tbsp tamari (or to taste)
        
        * Salt and pepper to taste
    
    Instructions:
        
        1. Tamari Glaze: In a small saucepan, combine almond butter, apple cider vinegar, agave nectar, and tamari over medium heat. Whisk continuously until smooth and well combined.
        
            - Bring the mixture to a simmer for about 5 minutes or until it thickens slightly.
        
        2. Pan-Seared Chicken: Season the chicken breast with salt and pepper on both sides.
        
            - Heat a skillet over medium-high heat with your preferred oil (olive or avocado work well).
            
            - Sear the chicken breasts for about 6 minutes per side or until cooked through.
            
        3. Kale Stir-Fry: Add more oil if needed in the same skillet used for cooking the chicken.
        
            - Add chopped kale to the skillet; cook for about 3 minutes or until wilted but still crisp.
            
        4. Assembly: Slice cooked chicken breasts into strips.
        
            - Combine sliced chicken with wilted kale in a serving dish.
            
            - Drizzle Tamari Glaze over top of both protein and greens.
        
    This recipe combines sweet from agave nectar with savory from tamari glaze while incorporating nutritious elements like kale and lean protein from seared chicken breast.
        
    Alternatively here are some other variations 
        
        (1) Pan-Seared Salmon With Kale Pesto \& Roasted Sweet Potatoes:
        
        For this meal you can also use salmon instead , this will provide extra omega rich foods which will support heart health
        
        (2)Stir Fried Kale W / Tamri-Glazed Salmon \& Apple Cider Vinaigrette
        
        Also provides an interesting combination of sweet \& tangy flavors .
        
        use sesame seeds garnish if desired

    \subboxtitle{NEW Third Response:}
    Based on the analysis of each Ki (Key Information) in relation to Q (Question), I have synthesized a new, enriched output that integrates the information contained in these 'Vi' to meet the demand.
    
    Recommended Dinner Recipe:
    
    After analyzing all four options, I conclude that 
    
    Pan-Seared Chicken and Kale Stir-Fry with Tamari Glaze from Option 1 scores highest amongst all. The recipe combines sweet from agave nectar with savory from tamari glaze while incorporating nutritious elements like kale and lean protein from seared chicken breast.
    
    Here's the integrated recipe:
    
    Pan-Seared Chicken and Kale Stir-Fry with Tamari Glaze
    
    Servings: 4
    
    Ingredients:
    
        - 1 lb boneless, skinless chicken breast
        
        - 2 cups kale, stems removed and chopped
        
        - 2 tbsp almond butter
        
        - 1 tsp apple cider vinegar
        
        - 1 tsp agave nectar
        
        - 2 tbsp tamari (or to taste)
        
        - Salt and pepper to taste
        
    Instructions:
        
        1. Tamari Glaze: In a small saucepan, combine almond butter, apple cider vinegar, agave nectar, and tamari over medium heat. Whisk continuously until smooth and well combined.
        
            - Bring the mixture to a simmer for about 5 minutes or until it thickens slightly.
            
        2. Pan-Seared Chicken: Season the chicken breast with salt and pepper on both sides.
        
            - Heat a skillet over medium-high heat with your preferred oil (olive or avocado work well).
            
            - Sear the chicken breasts for about 6 minutes per side or until cooked through.
        
        3. Kale Stir-Fry: Add more oil if needed in the same skillet used for cooking the chicken.
        
            - Add chopped kale to the skillet; cook for about 3 minutes or until wilted but still crisp.
            
        4. Assembly: Slice cooked chicken breasts into strips.
        
            - Combine sliced chicken with wilted kale in a serving dish.
            
            - Drizzle Tamari Glaze over top of both protein and greens.
            
    This recipe combines sweet flavors from agave nectar with savory notes from tamari glaze while adding nutritious elements like kale and lean protein from seared chicken breast.
        
    Nutritional benefits:
        
        * Lean protein-rich diet supports muscle growth \& maintenance 
            
        * Rich fiber content helps maintain healthy gut bacteria .

    \subboxtitle{NEW Forth Response:}
    Based on the relationships analyzed (AS:) by THE ATTENTION MECHANISM about the relationship between the text Q and each Ki in conjunction with the corresponding Vi, integrating the information contained in these 'Vi' to meet the demand, I synthesize a new, enriched output as follows:
    
    Dinner Recipe Recommendations Based on Available Ingredients
    
    After analyzing the relevance of each Key Ingredient (Ki) to creating a dinner recipe using available ingredients, we have identified three primary combinations that offer unique flavor profiles and nutritional benefits. Here are four dinner recipe recommendations that incorporate all given ingredients:
    
        1. Pan-Seared Chicken and Kale Stir-Fry with Tamari Glaze: This recipe combines sweet from agave nectar with savory from tamari glaze while incorporating nutritious elements like kale and lean protein from seared chicken breast.
        
        2. Pan-Seared Salmon with Kale Pesto and Roasted Sweet Potatoes: This dish combines protein-rich salmon with nutritious sweet potatoes and flavorful kale pesto made from available ingredients.
        
        3. Stir-Fried Kale with Tamari-Glazed Salmon and Apple Cider Vinaigrette: This meal combines sweet flavors from tamari glaze on salmon with tangy notes from apple cider vinaigrette while adding crunchy texture from sautéed kale.
        
        4. Pan-Seared Salmon with Kale and Apple Cider Glaze: This creative dish features a pan-seared salmon fillet topped with wilted kale, an apple cider glaze made by blending almond butter, apple cider vinegar, and agave nectar.
        
    To complement these recipes, consider preparing a side dish featuring tamari-glazed carrots for added depth of flavor.
    
    Common Themes Across Recipes:
        
        * Utilizing almond butter as a creamy base for sauces or pesto
        
        * Incorporating apple cider vinegar for its tangy flavor profile
        
        * Pairing tamari glazes or sauces to enhance savory flavors
        
        * Emphasizing nutrient-dense vegetables like kale for added health benefits
        
    These recipes showcase how combining seemingly disparate ingredients can result in innovative dishes that cater to diverse tastes while emphasizing healthy eating habits.
    
    Please note: These recipes are tailored based on analysis provided \& may need adjustments according to taste preferences

\end{examplebox}

\begin{examplebox}{Example 3}

    \subboxtitle{Query:}
    Compute: $1-2+3-4+5- \dots +99-100$.
    
    Please reason step by step, and put your final answer within \boxed{}.

    \subboxtitle{First Response:}
    To compute the given expression, let's break it down into smaller parts.

    The given expression is: $1-2+3-4+5-\dots +99-100$

    We can group the terms in pairs:

    $ (1 - 2) + (3 - 4) + (5 - 6) + \dots + (99 - 100)$

    Now, we can simplify each pair:

    $(-1) + (-1) + (-1) + \dots$

    Since there are a total of $\frac{100}{2}=50$ pairs, and each pair has a value of $(-1)$, we have:

    $\underbrace{(-1)+(-1)+\cdots+(-1)}_{50\text{ times}}$

    This simplifies to $-50$. 

    Therefore, the final answer is:
    
    $$\boxed{-50}$$

    \subboxtitle{Second Response:}
    We can compute the given expression by grouping the terms into pairs:
    
    $$(1-2) + (3-4) + (5-6) + \dots + (99-100).$$

    Now, let's simplify each pair of terms. 

    For example, $(1-2)$ is equal to $(-1)$ and $(3-4)$ equals $(-1)$, etc.

    So our expression becomes: 
    
    $$(-1)+ (-1) + (-1) \dots  -  (-1).$$

    We have a total of $50$ such pairs since we're adding or subtracting numbers from $1$ to $100$, where every two consecutive integers form one pair. Therefore, there are $\frac{100}{2} = 50$ such pairs.

    Hence our final answer is:
    
    \boxed{-50}.

    \subboxtitle{Third Response:}
    To compute the given expression, we can pair the terms as follows:

    $1-2 = -1$
    
    $3-4 = -1$
    
    ...
    
    $(99-100) = -1$

    Now, let's count the number of pairs. Since we have a total of 100 terms (from 1 to 100), and each pair has two terms, there are $\frac{100}{2}=50$ pairs.

    Each pair contributes $(-1)$ to the sum. Therefore, with $50$ pairs contributing $(-1)$ each time, we have:
    
    Sum = Number of pairs * (-1)
    = $50 * (-1)$
    = $\boxed{-50}$

    \subboxtitle{Forth Response:}
    To compute the sum $1-2+3-4+5-\dots +99-100$, let's break it down step by step.

    First, notice that each pair of consecutive terms in this series has a difference of -1. 

    For example:
    
    (1 - 2) = -1
    
    (3 - 4) = -1
    
    ...
        
    (-97 +98 )=-(-97)=97 
    
    So if we group all these pairs together and add them up we get: (-99)+...+(96)+(95)-94=-(89)

    \subboxtitle{AttentionScore:}
    To evaluate the relevance of each Ki (Key Insight) to Q (Question), I will assess their performance, accuracy, and coherence in providing a solution to the problem. 

    Ki 1:
    
        Relevance: High
        
        Weight: 0.8

        Accuracy: 100\%

        Performance: Excellent

    This Ki provides a clear step-by-step explanation of how to compute the given expression by grouping terms into pairs. The simplification process is accurate and easy to follow.

    Ki 2:
    
        Relevance: Medium-High
        
        Weight: 0.7
        
        Accuracy: 100\%
        
        Performance: Good

    Similar to Ki-1, this insight also groups terms into pairs but doesn't explicitly state that each pair contributes -1 to the sum.

    Ki 3 :

        Relevance : Medium-High 
        
        Weight :0.6 
        
        Accuracy :90\% 

    Although this key insight breaks down the steps correctly it makes an incorrect assumption about all term pairs contributing -1 as opposed too some having a different contribution . Although its still accurate ,it's not quite as good at explaining why we get -50 .

    Ki4: 
    
        Relevance:L ower 
        
        Weight:.4  
 
        Accuracy :60\%

    This key insight attempts to break down each pair's contribution but fails in accurately describing how they contribute .It incorrectly suggests that when subtracting two consecutive numbers from one another it results in adding positive number instead ,this has resulted with an error so far off from other insights.

    Considering these weights and assessments, I will integrate them into my response:

    The given expression can be computed using various approaches. A relevant method involves breaking down the series into smaller parts by pairing consecutive terms:

    $ (1 - 2) + (3 - 4) + (5 - 6) + \dots + (99 - 100)$

    As seen in Ki-1 and reinforced by other insights, we can simplify each pair:

    $(-1) + (-1) + (-1) + \dots$

    With a total of $\frac{100}{2}=50$ pairs contributing $(-1)$ each time, we have:

    $\underbrace{(-1)+(-1)+\cdots+(-1)}_{50\text{ times}}$

    This simplifies accurately due primarily thanks too KI- One ($\boxed{-50}$).

    Therefore,
    
    $$\boxed{-50}$$

    \subboxtitle{AggregateUpdate:}
    To compute the given expression, we can pair the terms as follows:

    $1-2 = -1$
    
    $3-4 = -1$
    
    ...
    
    $(99-100) = -1$

    Now, let's count the number of pairs. Since we have a total of 100 terms (from 1 to 100), and each pair has two terms, there are $\frac{100}{2}=50$ pairs.

    Each pair contributes $(-1)$ to the sum. Therefore, with $50$ pairs contributing $(-1)$ each time, we have:

    Sum = Number of pairs * (-1)
    = $50 * (-1)$
    = $\boxed{-50}$

    \subboxtitle{Textual Loss:}
    Strengths of the model's response:

        1.  The model correctly identifies the pattern in the given expression and pairs the terms.
        
        2.  It accurately counts the number of pairs and understands that each pair contributes -1 to the sum.
        
        3.  The final answer is calculated correctly using simple arithmetic.

    Weaknesses of the model's response:

        1.  There is no explanation for why pairing terms leads to a simpler calculation, or how this approach helps to identify patterns in similar expressions.
        
        2.  No discussion on potential misconceptions or alternative approaches (e.g., considering negative numbers) that might arise when working with paired terms.

    \subboxtitle{Textual Gradient:}
    Refined PS\_prompt (to instruct PS\_Model):

    "Please explain how pairing terms simplifies calculations in expressions like $1-2+3-4+5-\dots +99-100$. Also, describe any potential misconceptions or alternative approaches related to paired terms."

    This prompt encourages PS\_Model to provide more insight into its thought process and consider various perspectives on paired-term calculations.

    Refined PU\_prompt (to guide PU\_Model):

    "Given attention analysis results from PS\_Model highlighting key concepts such as term pairing, counting pairs, and contributions (-1 per pair), generate an answer by selecting relevant information from existing answers while ensuring alignment with these key concepts."

    This prompt enables PU\_Model to leverage attention analysis results effectively by emphasizing key concepts during answer generation.

    \subboxtitle{NEW First Response:}
    To compute the given expression, we can pair the terms as follows:
    
    $1-2 = -1$
    
    $3-4 = -1$
    
    ...
    
    $(99-100) = -1$
    
    By pairing these consecutive integers, we can simplify our calculation and make it easier to understand.
    
    Now, let's count the number of pairs. Since we have a total of 100 terms (from 1 to 100), and each pair has two terms, there are $\frac{100}{2}=50$ pairs.
    
    This is because when dividing an even number by 2 (the number of elements in a pair), we get an integer result that represents the total number of complete pairs. If there were an odd number of terms, however, adding a dummy term would still allow us to form complete pairs.
    
    Each pair contributes $(-1)$ to the sum. Therefore, with $50$ pairs contributing $(-1)$ each time based on our attention analysis results highlighting key concepts such as term pairing and counting pairs,
    
    we have:
    
    Sum = Number of pairs * (-1)
    = $50 * (-1)$
    = $\boxed{-50}$

    \subboxtitle{NEW Second Response:}
    To compute the given expression, we can pair the terms as follows:
    
    $1-2 = -1$
    
    $3-4 = -1$
    
    ...
    
    $(99-100) = -1$
    
    Now, let's count the number of pairs. Since we have a total of 100 terms (from 1 to 100), and each pair has two terms, there are $\frac{100}{2}=50$ pairs.
    
    It is worth noting that even if there were an odd number of terms, we could still pair them by adding a dummy term. For instance, if we had $101$, it would be paired with $0$. 
    
    Each pair contributes $(-1)$ to the sum. Therefore, with $50$ pairs contributing $(-1)$ each time (based on our attention analysis highlighting key concepts such as term pairing and counting pairs), we have:
    
    Sum = Number of pairs * (-1)
    = $50 * (-1)$
    = $\boxed{-50}$
    
    This result makes sense because it is equivalent to summing an arithmetic series with consecutive integers from 0 to 100. According to properties of arithmetic series, this sum would indeed equal half the product of its first and last term minus its last term ($\frac{n(a_0 + a_n)}{2} - a_n$, where n=101). Substituting in our values for this formula gives us: $(\frac{101*0}{2}-0)-(\frac{101*200}{2})=-5050+200/102=-5029-\boxed{-49}.$ This doesn't match out answer since instead using simply math notations here yields no discrepancy when obtaining answer through various methods.

    \subboxtitle{NEW Third Response:}
    To compute the given expression, we can pair the terms as follows:
    
    $1-2 = -1$
    
    $3-4 = -1$
    
    ...
    
    $(99-100) = -1$
    
    We should note that this approach works for any number of consecutive integers. If there are an odd number of terms, we can still pair them by adding a dummy term. For example, if we have 101 terms (from 1 to 101), we can add a dummy term (e.g., $0$) to make it even and then apply our pairing strategy.
    
    Now, let's count the total number of pairs. Since each pair has two terms and all integers from $1$ to $100$ are grouped consecutively in pairs except for one extra integer at either end when dealing with an odd-length sequence (in which case a "dummy" value is effectively added), there are $\frac{100}{2}=50$ pairs.
    
    Each pair contributes $(-1)$ to the sum because when you subtract consecutive numbers in arithmetic progression like this ($n-(n+2)$ where n starts at 3), you will always get a difference of $-2$ between those two numbers; hence every time such subtraction occurs within each "pair", it adds up another $-(-)$ or $+(+)$. Therefore with fifty-two total occurrences here due partly thanks again because last item isn't paired properly since no partner found thus giving us final result after removing 'dummy' added earlier so now only forty-nine remain left over giving thirty-one negative ones remaining plus twenty positive ones previously accounted leaving just ten on remaining list nothing else remains unaccounted leaving only nine items left including both endpoints themselves making sure they do indeed contribute correctly towards overall product finally yielding desired outcome. 
    
    Sum = Number of pairs * (-1)
    = $50 * (-1)$
    = $\boxed{-50}$

    \subboxtitle{NEW Forth Response:}
    To compute the given expression, we can pair the terms as follows:
    
    $1-2 = -1$
    
    $3-4 = -1$
    
    ...
    
    $(99-100) = -1$
    
    Now, let's count the number of pairs. Since we have a total of 100 terms (from 1 to 100), and each pair has two terms, there are $\frac{100}{2}=50$ pairs.
    
    To simplify this calculation, notice that pairing terms leads to an arithmetic series with consecutive integers. Specifically, if there were an odd number of terms, we could still pair them by adding a dummy term (e.g., $0 + n$, where $n$ is any integer). This approach allows us to exploit properties of arithmetic series.
    
    Each pair contributes (-1) to the sum. Therefore, with $50$ pairs contributing (-1) each time based on our attention analysis highlighting key concepts such as term pairing and counting pairs, we have:
    
    Sum = Number of pairs * (-1)
    = $50 * (-1)$
    = $\boxed{-50}$
    
    Note that our analysis is consistent with known results in mathematics regarding the sum of consecutive integers. By breaking down each step into separate sentences (First, let's identify pairs by grouping consecutive integers; Next, count the total number of pairs), this explanation should be easier to follow.

\end{examplebox}

\begin{examplebox}{Example 4}

    \subboxtitle{Query:}
    A banking company is successfully operating its public mobile banking stack on AWS. The mobile banking stack is deployed in a VPC that includes private subnets and public subnets. The company is using IPv4 networking and has not deployed or supported IPv6 in the environment. The company has decided to adopt a third-party service provider's API and must integrate the API with the existing environment. The service provider’s API requires the use of IPv6.
    
    A network engineer must turn on IPv6 connectivity for the existing workload that is deployed in a private subnet. The company does not want to permit IPv6 traffic from the public internet and mandates that the company's servers must initiate all IPv6 connectivity. The network engineer turns on IPv6 in the VPC and in the private subnets.
    
    Which solution will meet these requirements?

    A. Create an internet gateway and a NAT gateway in the VPC. Add a route to the existing subnet route tables to point IPv6 traffic to the NAT gateway.

    B. Create an internet gateway and a NAT instance in the VPC. Add a route to the existing subnet route tables to point IPv6 traffic to the NAT instance.

    C. Create an egress-only Internet gateway in the VPAdd a route to the existing subnet route tables to point IPv6 traffic to the egress-only internet gateway.

    D. Create an egress-only internet gateway in the VPC. Configure a security group that denies all inbound traffic. Associate the security group with the egress-only internet gateway.

    \subboxtitle{First Response:}
    The correct solution is D. Create an egress-only internet gateway in the VPC. Configure a security group that denies all inbound traffic. Associate the security group with the egress-only internet gateway.

    Explanation:
        
        - Egress-only Internet gateways allow IPv6 traffic to be sent out of a VPC but block incoming IPv6 traffic.

        - This meets the requirement that the company's servers must initiate all IPv6 connectivity and do not permit IPv6 traffic from the public internet.

        - By configuring a security group that denies all inbound traffic associated with the egress-only Internet gateway, we ensure that no incoming IPv6 traffic can reach any resources in your private subnets.

    Note: NAT gateways and instances are typically used for transitioning between private IP addresses and public IP addresses, which is usually done for transition between IPv4 networks (e.g., inside data center or AWS network vs. outside world). However, they don't support stateless translation which means they cannot translate random source port numbers as required by most modern protocols such as TCP over UDP or TCP/UDP over UDP which most services use today.

    Therefore option A and B are incorrect because NATs only work on IPV4 they do not support IPV6 stateful translations

    \subboxtitle{Second Response:}
    The correct solution is:

    C. Create an egress-only Internet gateway in the VPC. Add a route to the existing subnet route tables to point IPv6 traffic to the egress-only internet gateway.

    An egress-only internet gateway (EOIGW) allows outbound IPv6 traffic from your VPC but blocks inbound IPv6 traffic, which meets the requirement of not permitting IPv6 traffic from the public internet. The EOIGW only permits outgoing connectivity initiated by resources within your VPC, ensuring that your servers initiate all IPv6 connectivity as required.

    By adding a route in the existing subnet's route table to point IPv6 traffic to the EOIGW, you ensure that all outgoing IPv6 requests are sent through this gateway and meet both requirements: 

        - Only allow outgoing connections
        
        - Prevent inbound connections

    \subboxtitle{Third Response:}
    The correct solution is C: Create an egress-only Internet gateway in the VPC. Add a route to the existing subnet route tables to point IPv6 traffic to the egress-only internet gateway.

    Here's why:

        - The company wants IPv6 connectivity initiated from their servers, not from the public internet. An egress-only Internet gateway allows outgoing IPv6 traffic (egress) but blocks incoming IPv6 traffic (ingress). This meets the requirement of initiating all IPv6 connectivity from within the company's environment.
        
        - The use of an egress-only Internet gateway aligns with Amazon Web Services' best practices for implementing strict outbound security policies while still allowing necessary communication with third-party services.
        
        - Creating an NAT instance or NAT Gateway would allow incoming connections which contradicts one of your requirements.
  
    Creating a security group that denies all inbound traffic and associating it with an EGRESS ONLY GATEWAY will indeed block any external IP reachability into your resources - as shown in D, however this answer does not mention adding routes in subnets pointing towards this new resource; therefore option C is more fitting

    \subboxtitle{Forth Response:}
    The correct solution is:

    C. Create an egress-only Internet gateway in the VPC. Add a route to the existing subnet route tables to point IPv6 traffic to the egress-only internet gateway.

    Here's why:

        - The company wants to initiate all IPv6 connectivity from their servers, which means they don't need any inbound connections.
        
        - They also want to prevent any IPv6 traffic from entering their VPC from the public internet.
        
        - An egress-only Internet gateway (EOIGW) allows outgoing IPv6 traffic from instances within a VPC but blocks incoming connections. 
        
        - By creating an EOIGW and adding a route that points IPv6 traffic for all routes (or specific routes if needed) in the private subnets' routing tables, you can ensure that only outgoing ipv4 \& ipv6 traffic will flow out of this vpc without permitting incoming ipv4 or ipv6 connection attempts.

    Option A is incorrect because it uses a NAT Gateway which does allow both ingress and egress of data packets on behalf of resources within your network by hiding your internal IP addresses.

    \subboxtitle{AttentionScore:}
    To evaluate the relevance of Q to each Ki, I will break down the key characteristics of each Ki in terms of performance, accuracy, and coherence.

    Ki (1)

    * Key characteristics: 
        
        - Correct solution is D. Create an egress-only internet gateway in the VPC. Configure a security group that denies all inbound traffic. Associate the security group with the egress-only internet gateway.
        
        - Egress-only Internet gateways allow IPv6 traffic to be sent out of a VPC but block incoming IPv6 traffic.
        
        - This meets the requirement that the company's servers must initiate all IPv6 connectivity and do not permit IPv6 traffic from the public internet.

    * Relevance to Q: High
    
    * Why: The solution provided by Ki(1) aligns with Amazon Web Services' best practices for implementing strict outbound security policies while still allowing necessary communication with third-party services.

    Ki (2)

    * Key characteristics:
        
        - Correct solution is C. Create an egress-only Internet gateway in the VPC. Add a route to existing subnet route tables to point IPv6 traffic to this egress only Gateway
        
        - An egress-only internet gateway allows outgoing ipv4 \& ipv6 data packets but blocks any incoming connection attempts on behalf of resources within your network by hiding your internal IP addresses.
    
    * Relevance to Q: Medium
    
    * Why: Although this solution also uses an egress-only Internet gateway, it does not take into account creating a security group that denies all inbound traffic associated with it.

    Ki (3)

    * Key characteristics:
            
        - The company wants IPv6 connectivity initiated from their servers, not from public internet so using an EOIGW is suitable as it enables outgoing connections while blocking any incoming connections
            
        - Creating NAT instance or NAT Gateway would allow both ingress which contradicts one requirement hence isn't needed here,
	   
        + However since we are looking at IPV4 transitions then option A seems more applicable than B as you can’t transition between IPV4 private networks via stateless translation protocol used typically these days even though there aren’t mentioned specifically; although most modern protocols use random source port numbers however most services use TCP/UDP which makes statement incorrect anyway  
        
    Note: although using stateless translations would solve issue no ISP supports such technology today according standard rules based upon current day’s applications thus non functional without modification made previously done – its unclear whether further details were left missing either way regardless correct answer should ideally come up before deciding following through below instructions otherwise spend extra time doing research/analyzing background information needed beforehand thereby failing current given constraints set forth properly explained further underneath step-by-step process described prior coming back different aspect potential considerations another candidate chose overall observation lastly followed comments above contained inside somewhat confusing explanation regarding similar components involved service implementations due lower confidence level seen several presented arguments
    
   + Since our task involves integrating API requiring support for ipV64 therefore won't be going ahead instead concentrate solely understanding reasons behind decision choice made currently along required components completely utilized during procedure scenario result total proper resolution outline correctly identified fully later question turned recognized.*
   
   Relevance-to-Q: Low
   
   Why: Even though this response contains some accurate information about why other options might not work well for certain requirements like having random source ports etc., they lack sufficient detail providing full clarity surrounding entire situation potentially resulting incorrect answers depending readers interpretation specific technical context when handling possibly complex matters related implementating individual system choices identifying possible flaws recognize accordingly need additional details supplemental input clarify important issues omitted known misleading examples cited prior taken note focused discussion concentrated irrelevant factor representative considered alternative approached originally suggested much higher ranked found first evaluated strongly preferred appropriate simpler alternative proposed selecting avoid overcomplicating things picking easier whole operational working complete start over simply everything else falls under except final critical factor analysis indication where conclusive difference arises.

    Ki(4)

    Key-characteristics:

        - Company wants initiate all ipv4 \& ipv65 connectivity server-side avoiding external originating sources entry routing table modifications specify desired destinations modifying local subnets affecting primarily everywhere outside existing boundary meaning inward communications restricted taking both directions blocked besides mentioned respective property participating utilization negligible mainly reaching enforcing restrictions specifications outlined earlier specification addressed nearly equivalent primarily results stronger substantial policy compliance finding ensuring directed understanding throughout procedures shown original correctness maintained.

    Relevance-to-Q :High

    Why : This response takes into consideration multiple aspects including initiating server-side connections and preventing external origins from entering your network through stated procedure practiced others formerly presented differences arise due correctly referenced recent updates.

    Now let's integrate these insights seamlessly into one output:

    The correct solution is C. Create an egressive only Internet Gateway inside vpc adding route towards subnets pointing towards newly created resource.

    Reasoning :

        - Company wants initiate ALL OUTGOING ONLY Connections FROM THEIR RESOURCES OUTSIDE SO NO INCOMING ACCESS WANTED FOR ANY REASON BY DENYING INBOUND TRAFFIC ON SECURITY GROUP AS WELL

    Creating NAT instances or Gateways doesn’t fit because they allow BOTH Inbound AND Outbound Traffic whereas what wanted was just pure Outgoing initiated entirely without letting External Originators send anything back.

    Additionally AWS recommends using EGRESS ONLY GATEWAY WHEN STRONGEST SECURITY POLICY IS REQUIRED WITH MINIMAL COMMUNICATION BETWEEN SERVERS AND THIRD PARTY SERVICES WHILE STILL MEETING BUSINESS NEEDS.

    Summary Qualitative Description:

    To meet business needs enable servers initiates ALL OUTGOING CONNECTIONS block Incoming Traffic Initiated From Public Sources ensure strongest Security Policy Implemented Using Recommended Approach By AWS

    \subboxtitle{AggregateUpdate:}
    The correct solution to meet the requirements is:

    C. Create an egress-only Internet gateway in the VPC. Add a route to the existing subnet route tables to point IPv6 traffic to the egress-only internet gateway.

    Reasoning:

        - The company wants initiate all outgoing connections from their resources outside so no incoming access is wanted for any reason by denying inbound traffic on security group as well.
        
        - Creating NAT instances or Gateways doesn’t fit because they allow both inbound and outbound traffic whereas what wanted was just pure outgoing initiated entirely without letting external originators send anything back.
        
        - Additionally, AWS recommends using EGRESS ONLY GATEWAY when strongest security policy is required with minimal communication between servers and third-party services while still meeting business needs.

    This solution ensures that only outgoing IPv6 requests are sent through the egress-only internet gateway, meeting both requirements of initiating all IPv6 connectivity from within the company's environment and preventing any IPv6 traffic from entering their VPC from the public internet.

    Explanation:
    
    Egress-only Internet gateways (EOIGWs) allow outbound IPv6 traffic from your VPC but block incoming IPv6 traffic, which meets one requirement: not permitting IPv6 traffic from public networks. They also only permit outgoing connectivity initiated by resources within your VPC, ensuring that your servers initiate all necessary communications as required.

    By adding a route in existing subnets' routing tables pointing towards EOIGW you can control how ipv4 \& ipv65 data packets flow out of this vpc - making sure it follows desired course action taken.

    Here's a summary of qualitative description:

    To meet business needs enable servers initiates ALL OUTGOING CONNECTIONS block Incoming Traffic Initiated From Public Sources ensure strongest Security Policy Implemented Using Recommended Approach By AWS

    \subboxtitle{Textual Loss:}
    Model Evaluation

    Strengths:

        1. The model correctly identifies the need to prevent incoming IPv6 traffic from the public internet.
        
        2. It recognizes that egress-only Internet gateways (EOIGWs) are recommended by AWS for strong security policies.
        
        3. The model understands that EOIGWs allow only outgoing connectivity initiated by resources within the VPC.

    Weaknesses:

        1. The model's explanation lacks clarity on why NAT instances or gateways are not suitable solutions.
        
        2. It does not explicitly mention that egress-only Internet gateways block incoming IPv6 traffic, which is a crucial requirement.
        
        3. Some sentences contain redundant information or awkward phrasing.

    \subboxtitle{Textual Gradient:}
    Refined Prompt for PS\_Model

    To better align attention and focus on specific aspects, please refine your prompt as follows:

    "Given a banking company's mobile banking stack deployed in a VPC with private and public subnets using IPv4 networking, they want to integrate a third-party service provider's API requiring IPv6 connectivity while ensuring servers initiate all outgoing connections without permitting inbound traffic from public sources.

    Please provide a solution to meet these requirements, explaining why other options (e.g., NAT instances/gateways) do not fit."

    This revised prompt emphasizes the importance of blocking incoming traffic from public sources and initiating all outgoing connections while highlighting potential pitfalls with alternative solutions.

    Refined Prompt for PU\_Model

    To guide the PU\_Model on utilizing attention analysis results from the PS\_Model to generate improved answers, please revise your prompt as follows:

    "Based on [PS Model Output], analyze their reasoning about why certain solutions (e.g., NAT instances/gateways) do not meet business needs while others (e.g., egress-only Internet gateways) do.

    Using this understanding of what works and what doesn't, suggest an even more effective solution that prioritizes business requirements like server-initiated outbound connections and strongest security policies."

    This refined prompt encourages you to leverage insights gained from analyzing previous responses and explore new ideas for meeting customer needs more effectively.

    \subboxtitle{NEW First Response:}
    The correct solution to meet the requirements is:
    
    C. Create an egress-only Internet gateway in the VPC. Add a route to existing subnet route tables pointing towards EOIGW, which blocks incoming IPv6 traffic while ensuring servers initiate all necessary connections.
    
    This option meets both business needs by preventing any IPv6 traffic from entering their VPC from public networks and allowing only outgoing connectivity initiated by resources within their environment.
    
    Explanation:
    
    Egress-only Internet gateways (EOIGWs) are recommended by AWS for strong security policies because they allow only outgoing connectivity initiated by resources within the VPC, while blocking incoming IPv6 traffic. This approach ensures that servers initiate all necessary connections without permitting inbound traffic from public sources.
    
    By using an egress-only internet gateway, you can control how IPv4 and IPv6 data packets flow out of this VPC, ensuring that it follows the desired course of action taken. In contrast, NAT instances or gateways do not fit this scenario because they allow both inbound and outbound traffic, which does not meet business requirements.

    \subboxtitle{NEW Second Response:}
    The correct solution to meet the requirements is:
    
    C. Create an egress-only Internet gateway in the VPC and add a route to existing subnet route tables pointing towards EOIGW, which blocks incoming IPv6 traffic while ensuring servers initiate all necessary connections.
    
    This option stands out as it restricts incoming IPv6 traffic from public sources, aligning with business needs. In contrast, NAT instances or gateways allow both inbound and outbound traffic, making them unsuitable for this scenario. By utilizing an egress-only internet gateway, businesses can maintain strong security policies while meeting their connectivity requirements.
    
    Egress-only Internet gateways (EOIGWs) only permit outgoing connectivity initiated by resources within the VPC. This design ensures that servers inside the VPC control all necessary communications without allowing external originators to send data back into the environment.
    
    By selecting this option (C), companies can adhere to AWS's recommended approach for implementing strongest security policies while prioritizing server-initiated outbound connections.

    \subboxtitle{NEW Third Response:}
    The correct solution to meet the requirements is:
    
    C. Create an egress-only Internet gateway in the VPC. Add a route to existing subnet route tables pointing towards EOIGW, which blocks incoming IPv6 traffic while ensuring servers initiate all necessary connections.
    
    This solution meets business needs by restricting incoming IPv6 traffic and allowing outgoing connectivity initiated by resources within the VPC. Egress-only Internet gateways are recommended by AWS for strong security policies as they only permit outgoing connectivity initiated by resources within the VPC, thus preventing any IPv6 traffic from entering their VPC from public networks.
    
    In contrast, NAT instances or gateways allow both inbound and outbound traffic, making them unsuitable for this scenario where we need to restrict incoming connections while still enabling server-initiated communications with third-party services over IPv6.
    
    By implementing option C, we can ensure that our servers initiate all necessary connections without permitting inbound traffic from public sources, thereby maintaining strongest security policies using a recommended approach by AWS.

    \subboxtitle{NEW Forth Response:}
    The correct solution to meet the requirements is:
    
    C. Create an egress-only Internet gateway in the VPC and add a route to existing subnet route tables pointing towards EOIGW, which blocks incoming IPv6 traffic while ensuring servers initiate all necessary connections.
    
    Reasoning:
        
        - The company requires that their servers initiate all outgoing connections without permitting inbound traffic from public sources.
        
        - Egress-only Internet gateways (EOIGWs) are recommended by AWS for strong security policies because they allow only outgoing connectivity initiated by resources within the VPC, while blocking incoming IPv6 traffic.
        
        - NAT instances or gateways do not fit this requirement because they allow both inbound and outbound traffic, whereas EOIGWs restrict incoming communications.
        
    This solution ensures that only outgoing IPv6 requests are sent through the egress-only internet gateway, meeting both requirements of initiating all IPv6 connectivity from within the company's environment and preventing any IPv6 traffic from entering their VPC from the public internet.
    
    Explanation:
    
    Egress-only Internet gateways block incoming IPv6 traffic, allowing our servers to initiate all necessary connections without external influence. This aligns with our business needs and adheres to AWS recommendations for robust security policies.

\end{examplebox}

\begin{examplebox}{Example 5}

    \subboxtitle{Query:}
    Write a riddle for children

    \subboxtitle{First Response:}
    Here's a fun riddle for kids:

    I have a face and two hands,
    
    But I'm not alive, that's not my plan.
    
    You can find me in the morning light,
    
    And I help you tell time, day or night.

    What am I?

    (Let me know if you want a hint or the answer)

    \subboxtitle{Second Response:}
    Here's a fun riddle for kids:

    I have a face and two hands,
    
    But I'm not alive, I don't stand.
    
    You can see me in the morning light,
    
    And I show you the time, day and night.

    What am I?

    (Let me know if you want any hints or answers)

    \subboxtitle{Third Response:}
    Here's a fun riddle for kids:

    I have a face and two hands,
    
    But I'm not alive, that's the plan.
    
    You can find me in the morning light,
    
    And I help you tell what time is right.

    What am I?

    \subboxtitle{Third Response:}
    Here's a fun riddle for kids:

    I have a face and two hands,
    
    But I'm not alive in this land.
    
    I show you the time with a grin,
    
    And help you know when to eat within.
    
    What am I?

    \subboxtitle{AttentionScore:}
    To evaluate the relevance of each Ki to Q, I will follow these steps:

    1. Compare the given Q with each Ki.
    
    2. Identify key characteristics of each Ki (performance, accuracy, etc.) in terms of relevance to Q.
    
    3. Explain why each Ki is relevant or not relevant to Q.
    
    4. Assign weights based on relevance and integrate insights.

    Ki 1:
    
        Q: "Write a riddle for children"
    
        Ki 1: 
    
        "I have a face and two hands, But I'm not alive, that's not my plan."

    Comparison:
    
    Both share similar structures ("I have a face and two hands") but differ slightly in their second lines ("not my plan" vs "don't stand"). However both are focused on describing an object that is 'inanimate'.

    Relevance Score: High
    
    Weighted Value: 0.8

    Key Features: This riddle demonstrates good performance by following the typical structure expected from such riddles for kids while also maintaining its creative aspect through its choice of words like 'plan.'

    Ki 2:
    
        Q: "Write a riddle for children"
    
        Ki 2 : 
    
        "I have a face and two hands,
    
        But I'm not alive, I don't stand.
    
        You can see me in the morning light,
    
        And I show you the time day and night."

    Comparison:
    
    This shares more similarities with Ki-1 than differences as it follows almost identical patterns.

    Relevance Score :High 
    
    Weighted Value :0.85

    Key Features: Performance-wise this maintains high marks due to coherence between lines yet there seems less creativity used compared to other KIs which might decrease overall accuracy scores if assessing solely based on data output alone without further context information provided about intended answers; hence requiring weighted values.

    Ki-3
    
        Q:"Write a riddle for children"
        
        KI -3 :
        
        "I have a face \& Two Hands,
        
        But i'm Not Alive ,That's The Plan."
        
        You can find me In Morning Light \&
        
        Help You Tell What Time Is Right"

    Comparison:
        
    While comparing all three we notice slight variations including correct grammar use although unusual phrasing choices remain apparent here ('Not Alive' instead using conjunction).However context remains same throughout; indeed talking about non-living entity showing time 

    Relevance Score :High 
        
    Weighted Value :0.9

    Key Features: This particular answer does show promise when evaluated under our criterion since it incorporates proper syntax even though wordings seem somewhat forced at times potentially affecting performance.

    K-4
        
        Q:" Write A Riddle For Children "
        
        KI -4 :
        
        "i Have Face \& Two Hands ,
        
        But i'M Not Alive In this land ."

    Comparison :
    
    Another variation emerges again where only minor difference arises between sentences except last one involving location change ; however still conveys idea effectively enough 

    Relevance Score   High 
    
    Weighted value    0 .85

    Summary

    Based upon these observations above summary could be stated as below :

    In order create engaging yet simple puzzle suitable kids ,we must keep various factors into consideration.Firstly clear concise language plays important role followed closely behind creative yet unambiguous descriptions employed throughout narrative.Phrasing should ideally avoid sounding too artificial so as maintain attention focus towards central idea being conveyed.Based above evaluations especially noticeable trends emerge amongst presented responses showcasing certain levels proficiency shown within them despite few areas room improvement noted down respectively Overall goal remains deliver coherent fun challenge young minds willing embrace world imagination!

    \subboxtitle{AggregateUpdate:}
    Based on the analysis of each Ki and their weighted values, I will synthesize a new output that meets the demand for creating an engaging yet simple riddle suitable for children.

    Synthesized Output:

    Here's a fun riddle for kids:

    I have a face and two hands,
    
    But I'm not alive, that's my plan.
    
    You can find me in morning light bright,
    
    And I help you tell time, day or night.

    What am I?

    This synthesized output combines key features from various KIs while minimizing artificial phrasing and maintaining coherence throughout. It follows a clear structure similar to Ki-1 but incorporates creative aspects seen in other KIs like proper syntax (Ki-3) and concise language (Ki-2).

    Integration Insights:

    From these evaluations, we observe several notable trends among presented responses. Firstly, clear concise language plays an essential role in delivering engaging puzzles suitable for kids. Secondly, creative yet unambiguous descriptions are crucial to maintaining attention focus towards central ideas being conveyed.

    To improve upon existing answers we see certain areas room improvement noted respectively such as avoiding sounding too artificial so maintain attention focus towards central idea being conveyed.

    The synthesized output maintains high marks under our criterion since it effectively employs proper syntax even though wordings seem somewhat adapted at times potentially affecting performance due lack further information provided about intended answer choices.

    In order create engaging yet simple puzzle suitable kids ,we must keep following factors into consideration.Firstly clear concise language plays important role followed closely behind creative yet unambiguous descriptions employed throughout narrative.Phrasing should ideally avoid sounding too artificial so as maintain attention focus towards central idea being conveyed.Based above evaluations especially noticeable trends emerge amongst presented responses showcasing certain levels proficiency shown within them despite few areas room improvement noted down respectively Overall goal remains deliver coherent fun challenge young minds willing embrace world imagination!

    \subboxtitle{Textual Loss:}
    Model Evaluation:

    Strengths:
    
        1.  The synthesized output effectively combines key features from various KIs, showcasing the model's ability to integrate different aspects of language generation.
        
        2.  The riddle itself is engaging, simple, and suitable for children, adhering to the requested tone and style.
        
        3.  The analysis highlights notable trends among presented responses, demonstrating a deeper understanding of what makes an effective puzzle.

    Weaknesses:
        
        1.  While the output is coherent and engaging, it may benefit from more original phrasing to avoid sounding too adapted or artificial.
        
        2.  The integration insights could be more specific in identifying areas for improvement in existing answers rather than generalizing trends.

    \subboxtitle{Textual Gradient:}
    Refined PS\_prompt:

    To better align attention towards certain aspects and refine the generated output:

    "Create a fun riddle suitable for kids that incorporates clear concise language (Ki-2) while maintaining creative yet unambiguous descriptions (Ki-4). Ensure phrasing sounds natural and avoids appearing too artificial (Ki-5). Focus on delivering an engaging yet simple puzzle that encourages young minds to explore their imagination."

    This revised prompt emphasizes specific key indicators (KIs) relevant to generating high-quality content tailored for children.

    Refined PU\_prompt:

    To guide PU\_Model on utilizing attention analysis results from PS\_Model:

    "Given attention weights assigned by PS\_Model indicating strong emphasis on Ki-2 ('clear concise language') and Ki-4 ('creative yet unambiguous descriptions'), generate a new set of possible answer options focusing on simplicity without sacrificing creativity or clarity. Utilize these weighted values as guidance when selecting words/phrases with optimal readability scores within your chosen range."

    By incorporating specific KI weights into this prompt, we direct PU\_Model towards analyzing which phrases are most likely preferred by users based on their linguistic preferences related clearness complexity balance present solution exhibits thus ensuring proper application respective context generation

    \subboxtitle{NEW First Response:}
    To synthesize a new, enriched output based on the relationships analyzed by the Attention Mechanism, I will integrate the information contained in 'Vi' to meet your demand.
    
    Enriched Output:
    
    A fun riddle for kids that combines elements from highly relevant Ki 1 and Ki 2, while minimizing deviations from other options like Ki 3 and Ki4. Here's an example:
    
    "I have a face and two hands,
    
    But I'm not alive in this land.
    
    I show you the time with a grin,
    
    And help you know when to eat within."
    
    This enriched output maintains high similarity with Q (riddle for children) as it retains the same structure and phrasing of other options. It also scores well on clarity and coherence due to its straightforward wording. Additionally, it has moderate creativity and uniqueness since it deviates slightly from others but still follows a common theme.
    
    By combining these characteristics, we can provide an engaging riddle for children that meets their demands while incorporating key features of highly relevant KIs.
    
    Performance Metrics:
    
        - Time taken to solve the riddle: Fast
        
        - Accuracy in solving the riddle: High
        
    The synthesized riddle inherits performance metrics from highly relevant KIs like Ki 1, ensuring that children can efficiently solve it with high accuracy.
    
    \subboxtitle{NEW Second Response:}
    To synthesize a new, enriched output based on the relationships analyzed by the Attention Mechanism about the relationship between Q and each Ki in conjunction with the corresponding Vi, I will integrate information from relevant Ki's.
    
    Based on relevance weights and features:
    
        - Ki-1 is highly relevant to Q with high performance and accuracy.
        
        - Ki-2 is extremely relevant to Q with very high performance and accuracy. It shares identical phrase structure ("I have a face and two hands") but has slight variation in wording compared to Ki-1.
        
        - Therefore I can use this as base for my riddle
        
        - "I have a face and two hands,
        
        But I'm not alive, that's not my plan.
        
        You can find me in the morning light,
        
        And I help you tell time, day or night."
        
        Since it is already complete then no need further integration of any other kis
    
    \subboxtitle{NEW Third Response:}
    Based on the relationships analyzed by THE ATTENTION MECHANISM, I will synthesize a new enriched output that combines features observed across different examples.
    
    Here's an attempt at writing a new coherent text integrating insights from these insights seamlessly:
    
    "For this task, we can use fun kid-friendly riddles about objects showing time clearly. One possible approach is to describe these objects using common phrases often used when talking about watches or clocks. These riddles typically involve clever wordplay and subtle hints that help kids guess the answer.
    
    Let's analyze some characteristics of these riddles. Firstly, they all describe an object with a face and two hands, but emphasize that it's not alive. This suggests that the object is inanimate and helps us tell time.
    
    Secondly, all these riddles mention being found in morning light, which implies visibility during daylight hours. Additionally, they highlight the importance of telling time accurately throughout the day (day or night/day and night).
    
    Lastly, each of these examples shows how creative language can be used to conceal yet reveal answers simultaneously - for example using metaphors ("spinning") or making puns ("plan").
    
    With this understanding in mind here are four potential solutions based purely upon principles stated above: 
        
        1."I'm round like earth,
        
        But my spinning isn't worth,
        
        My hands move forward slowly you see,
        
        I show you hours passing every day."
        
        2."I have a face so bright,
        
        Two hands moving swiftly through life tonight.
        
        You find me everywhere where people reside,
        
        And I assist you plan your lives inside."
        
        3."Imagine something never still 
        
        yet marking every single moment chill 
        
        Helping make plans indeed very neat
        
        With my circular clock  watch not seen unseated"
        
        4."There are many shapes however some hide their movements slow 
        
        Then there’s one with only what moves though
        
        My way may look deceiving tricky but real sight exists always true  
        
        Which reveals exact details hour by hour right."
        
        (Note: The new generated solution contains elements similar to those described in V1-V4)

    \subboxtitle{NEW Forth Response:}
    Based on the analysis and key features observed, I can synthesize a new, enriched output that meets the demand.
    
    Enriched Riddle for Children:
    
    Here's a fun riddle for kids:
    
    I'm not alive, but I'm always bright,
    
    In morning light, I shine with delight.
    
    I have a face and two hands to show,
    
    The time of day or night, don't you know?
    
    What am I?
    
    This enriched riddle incorporates elements from all four original Ki options while maintaining the same tone and style. It includes clues such as being "always bright" in morning light, having a "face" and "two hands," and showing the time of day or night. The use of rhyme and meter adds to its fun and engaging nature for children.
    
    Key Features:
    
        1. Relevance: This enriched riddle maintains an extremely high relevance score (around 95\%) compared to the original question.
        
        2. Accuracy: The answer is still likely to be interpreted literally as a clock.
        
        3. Engagement: The use of rhyme, meter, and imaginative language makes it appealing to children's sense of wonder.
        
        V: (Enriched Riddle)
        
        This response has been updated with improvements based on given information regarding characteristics shared by different Ki items \& correspondingly added more engaging phrases into rewritten version so make it even more better suited according needs specified initially!
        
\end{examplebox}

\end{document}